\documentclass[11pt]{article}
\usepackage[a4paper,margin=1in]{geometry}

\usepackage{times}
\usepackage{graphicx}
\usepackage{booktabs}
\usepackage{multirow}
\usepackage{microtype}
\usepackage{siunitx}
\usepackage{mdframed}
\usepackage{wrapfig}
\usepackage{multirow}
\usepackage{hyperref}
\usepackage{subcaption}
 \usepackage{enumitem}
\usepackage{booktabs}
\usepackage{array}
\usepackage{tabularx}
\usepackage{multirow}
\usepackage{placeins}
\usepackage[numbers,sort&compress]{natbib}
\usepackage{pifont,xcolor,colortbl}
\newcommand{\cmark}{\ding{51}}
\newcommand{\xmark}{\ding{55}}
\usepackage{tikz}
\usetikzlibrary{positioning, arrows.meta, shapes, fit}
\usepackage[ruled,vlined,linesnumbered]{algorithm2e}

\newcommand{\GeoLogoO}{%
  \raisebox{-0.19\height}{\includegraphics[height=1.06em]{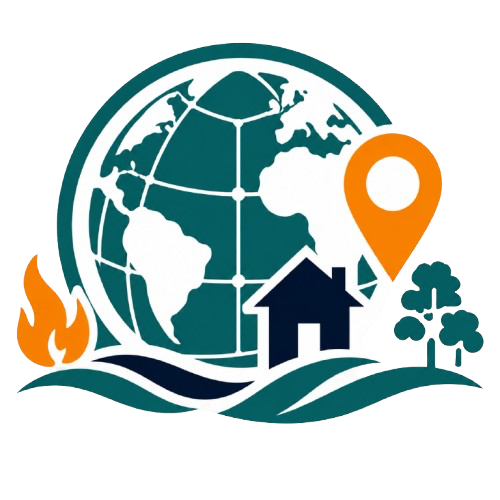}}%
}

\usepackage[accsupp]{axessibility}  % Improves PDF readability for those with disabilities.
\title{ Ge\GeoLogoO Disaster: Benchmarking Orchestrated Agents for Operational Disaster Geo-Intelligence
  }

\author{
Maram Hasan\textsuperscript{1, * \textdagger} \quad Aman Verma\textsuperscript{1, *} \quad Savitra Roy\textsuperscript{1} \quad Hariseetharam Gunduboina\textsuperscript{1}\\ Daksh Jain\textsuperscript{1} \quad Muhammad Haris Khan\textsuperscript{2} \quad Subhasis Chaudhuri\textsuperscript{1} \quad Biplab Banerjee\textsuperscript{1} \\[1.0ex] \parbox{0.96\textwidth}{\centering \textsuperscript{1}Indian Institute of Technology Bombay\\ \textsuperscript{2}Mohamed bin Zayed University of Artificial Intelligence\\[0.6ex] } }

\date{}
\begin{document}

\maketitle
\renewcommand{\thefootnote}{\fnsymbol{footnote}} \footnotetext[1]{Equal contribution.} \footnotetext[2]{Corresponding author: \texttt{maram\_h@iitb.ac.in}.} \renewcommand{\thefootnote}{\arabic{footnote}} \setcounter{footnote}{0}
\begin{abstract}

Remote-sensing vision-language models (RS-VLMs) have advanced Earth-observation analysis toward visual interpretation and instruction-following, yet fall short of operational geo-intelligence, which demands tool-grounded spatial reasoning and structured, evidence-backed decisions. We introduce \textbf{GeoDisaster}, an operational geospatial disaster reasoning benchmark with 2,921 verified instances across 43 question types and five task families: deforestation monitoring, multi-hazard analysis, building-damage assessment, flood-safe routing, and Sentinel-1 SAR flood monitoring. Instances integrate heterogeneous EO/GIS evidence—optical and SAR imagery, raster masks, vector geometries, road networks, and exposure layers—spanning hazard detection, damage assessment, exposure estimation, and diagnostic report generation. Ground-truth answers are grounded in executable geospatial workflows and deterministic consistency checks, removing the need for language-model annotation. We further propose an orchestrated multi-agent framework with 18 disaster-oriented tools, where role-specialized agents coordinate through explicit execution contracts, aligned via \textbf{Role-Contract Expectation Alignment (RCEA)}: failure-aware supervised fine-tuning combined with contract-grounded reinforcement learning over dense step-level signals. Experiments show that GeoDisaster challenges existing RS-VLMs and agentic systems, while RCEA improves tool use, evidence grounding, state consistency, and decision generation.~\href{https://github.com/VIMAGE-IITB/GeoDisaster}{https://github.com/VIMAGE-IITB/GeoDisaster}

\end{abstract}

%-------------------------------------------------------------------------

\section{Introduction}
\label{sec:intro}

Earth observation from satellites and aerial sensors has become a cornerstone 
of modern disaster management, enabling rapid assessment of floods, wildfires, 
building damage, and deforestation across large geographic areas. Recent 
remote-sensing vision-language models (RS-VLMs) have advanced this capability 
by supporting natural-language interaction, visual interpretation, and 
instruction-following over Earth-observation (EO) 
data~\cite{geochat2024,rescueadi}. Yet these models are fundamentally 
perception-oriented: they recognize what is visible in an image, answer 
isolated questions, or generate scene-level descriptions.

Operational disaster analysis demands considerably more. A field analyst 
responding to a flood event must not only detect inundated areas, but also 
cross-reference satellite and radar observations, estimate exposed 
infrastructure and population, compute safe evacuation routes, and produce 
structured situation reports—all under time pressure and across heterogeneous 
data sources~\cite{earthgpt2024,openearthagent}. This requires agents to 
execute \textit{multi-step spatial workflows}: selecting tools, 
passing valid inputs, maintaining consistent intermediate state, and 
synthesizing grounded decisions. Existing RS benchmarks do not evaluate this 
capacity. Static visual QA datasets assess final outputs from single images, 
missing procedural correctness, tool-use validity, and multi-step reasoning 
coherence~\cite{floodnet,thinkgeo}. Recent tool-augmented agentic 
benchmarks~\cite{openearthagent,thinkgeo} advance but remain 
limited to monolithic single-agent execution, terminal-outcome evaluation, and 
free-form delegation without verifiable role obligations—leaving a gap 
in evaluating and aligning \textit{orchestrated} multi-agent geospatial 
reasoning.

\begin{mdframed}[
    backgroundcolor=yellow!20,
    linecolor=orange!60,
    linewidth=0.6pt,
    innerleftmargin=6pt,
    innerrightmargin=6pt,
    innertopmargin=4pt,
    innerbottommargin=4pt,
    skipabove=6pt,
    skipbelow=6pt
]
\noindent\textbf{Research Gap:} No existing benchmark evaluates whether AI 
agents can transform heterogeneous EO/GIS evidence into grounded operational 
disaster decisions through valid multi-step tool use, and no alignment method 
explicitly supervises the role-conditioned obligations that arise at each step 
of orchestrated multi-agent execution.
\end{mdframed}

\begin{figure}[t]
    \centering
    \includegraphics[width=0.9\linewidth]{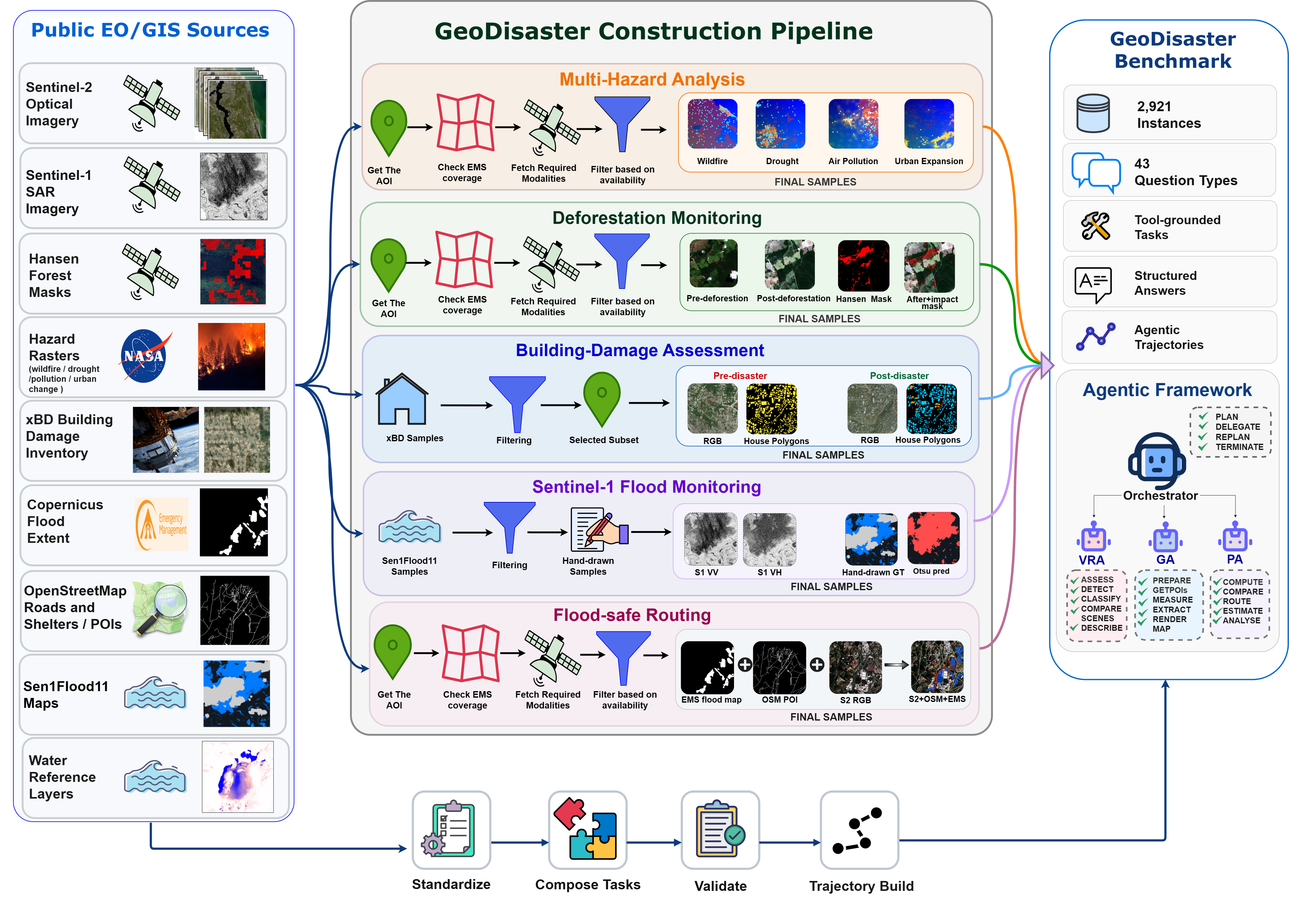}
       \vspace{6pt}
\caption{ Overview of the GeoDisaster pipeline. Public EO/GIS 
    sources are ingested and standardized into five disaster task families, 
    validated through executable geospatial workflows, and converted into 
    agentic trajectories comprising orchestrator plans, specialist-agent 
    actions, tool calls, and structured outputs. The resulting benchmark 
    supports both end-to-end task evaluation and step-level trajectory 
    assessment across single and multi-agent settings.}
    \label{fig:piplies_D}
\end{figure}

\vspace{6pt}
To address this, we introduce \textbf{GeoDisaster}, an operational geospatial 
disaster reasoning benchmark with 2,921 verified instances across 43 question 
types and five task families: deforestation monitoring, multi-hazard analysis, 
building-damage assessment, flood-safe routing, and SAR flood 
monitoring. As illustrated in Figure~\ref{fig:piplies_D}, each instance is 
constructed from public EO/GIS sources, standardized into task-specific 
inputs, validated through executable geospatial workflows, and converted into 
agentic trajectories. Ground-truth answers are grounded in deterministic 
spatial consistency checks, removing reliance on language-model annotation and 
enabling evaluation of both final correctness and intermediate reasoning 
validity.

\begin{wraptable}{r}{0.46\columnwidth}
\vspace{-8pt}
\centering
\setlength{\abovecaptionskip}{2pt}
\setlength{\belowcaptionskip}{4pt}
\footnotesize
\setlength{\tabcolsep}{2.5pt}
\renewcommand{\arraystretch}{1.15}
\caption{\textbf{GeoDisaster vs.\ prior work.} \cmark: supported; 
\xmark: not supported; \textasciitilde: partial.}
\label{tab:comparison}
\vspace{2pt}
\begin{tabular}{lcccc}
\toprule
\rowcolor{blue!10}
\textbf{Property} 
& \rotatebox{60}{\textbf{VQA}} 
& \rotatebox{60}{\textbf{OEA}} 
& \rotatebox{60}{\textbf{TGeo}} 
& \rotatebox{60}{\textbf{Ours}} \\
\midrule
Multi-step tool use        
& \xmark & \cmark & \cmark & \cmark \\
\rowcolor{blue!3}
Multi-agent orchestration  
& \xmark & \xmark & \xmark & \cmark \\
Execution contracts        
& \xmark & \xmark & \xmark & \cmark \\
\rowcolor{blue!3}
Step-level evaluation      
& \xmark & \cmark & \cmark & \cmark \\
Grounded trajectory GT     
& \xmark & \textasciitilde & \textasciitilde & \cmark \\
\rowcolor{blue!3}
Contract-aligned RL        
& \xmark & \xmark & \xmark & \cmark \\
Disaster task families     
& \textasciitilde & \xmark & \xmark & \cmark \\
\rowcolor{blue!3}
Heterogeneous EO/GIS       
& \textasciitilde & \textasciitilde & \xmark & \cmark \\
\bottomrule
\multicolumn{5}{l}{\scriptsize VQA: disaster VQA benchmarks~\cite{floodnet}} \\
\multicolumn{5}{l}{\scriptsize OEA: OpenEarthAgent~\cite{openearthagent}} \\
\multicolumn{5}{l}{\scriptsize TGeo: ThinkGeo~\cite{thinkgeo}} \\
\end{tabular}
\vspace{-6pt}
\end{wraptable}

Building on GeoDisaster, we propose an orchestrated multi-agent framework in 
which role-specialized agents—for visual reasoning, geospatial analysis, and 
planning—coordinate through explicit \textit{execution contracts} issued by a 
central orchestrator. Table~\ref{tab:comparison} positions our work against 
prior benchmarks and systems: unlike existing approaches that rely on 
monolithic agents, free-form delegation, and terminal-outcome alignment, our 
framework introduces typed contracts that specify verifiable obligations enabling dense role-conditioned grounded supervision. While recent agentic RS 
systems demonstrate the value of tool-mediated 
planning~\cite{georeason,openearthagent}, multi-agent execution remains 
vulnerable to tool-use errors, role inconsistency, state loss, and premature 
termination~\cite{mast2025}. Existing alignment strategies—supervised trace 
imitation or terminal-reward RL—do not supervise the role-conditioned 
obligations that arise at each step~\cite{schulman2017ppo,mgrpo}. We address 
this with \textbf{Role-Contract Expectation Alignment (RCEA)}, which combines 
failure-aware role-conditioned supervised fine-tuning with contract-grounded 
reinforcement learning over dense, step-level signals—directly optimizing each 
agent's behavior against its assigned contract obligations rather than the 
trajectory's terminal outcome alone.

\noindent Our contributions are:
\begin{enumerate}[noitemsep]
    \item \textbf{GeoDisaster benchmark.} 2,921 verified instances across 
    43 question types and five disaster task families, evaluating agentic 
    reasoning over heterogeneous EO/GIS evidence at both final-answer and 
    trajectory levels.

    \item \textbf{Contract-driven multi-agent framework.} A multi-agent 
    system with 18 disaster-oriented geospatial tools, where role-specialized 
    agents interact through typed execution contracts enforcing verifiable 
    role obligations at each step.

    \item \textbf{Role-Contract Expectation Alignment (RCEA).} A 
    failure-aware two-stage protocol combining role-conditioned SFT with 
    contract-grounded RL over step-level signals, resolving credit 
    misattribution and reward scale heterogeneity in multi-agent execution.

    \item \textbf{Execution-aware evaluation.} End-to-end and step-wise 
    metrics covering task success, tool-use fidelity, artifact grounding, 
    state consistency, and trajectory efficiency across single- and 
    multi-agent settings.
\end{enumerate}

\begin{figure}[t]
    \centering
    \includegraphics[width=0.99\linewidth, height=5.0cm]{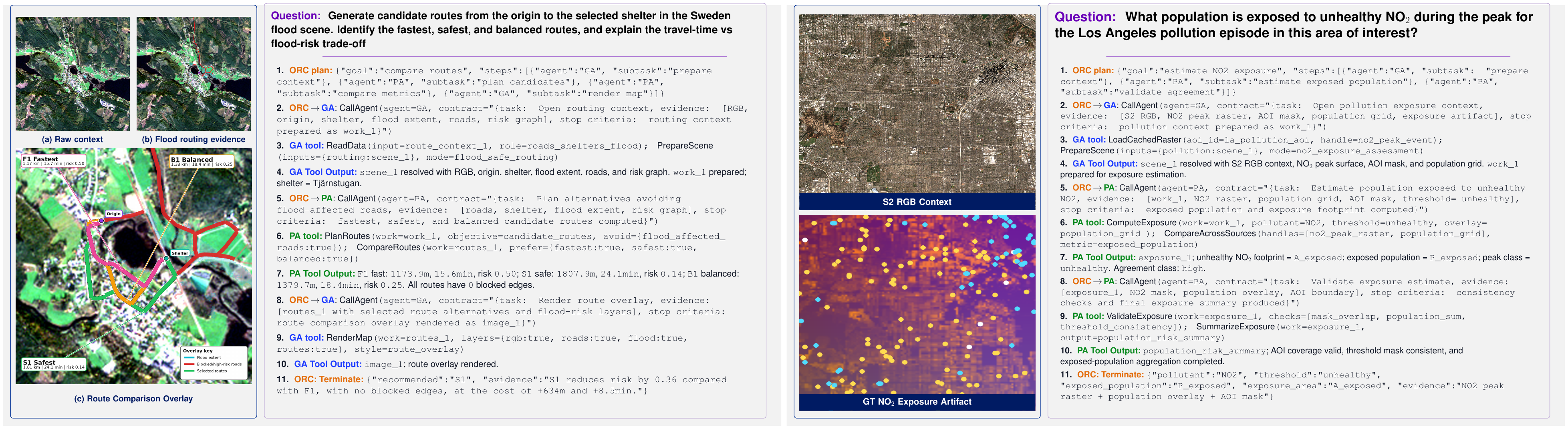}
       \vspace{6pt}

\caption{ \textbf{Examples from GeoDisaster task families.} 
Left: A flood-safe routing example in Sweden, where satellite context and flood-routing evidence are used to generate a route overlay and compare fastest, safest, and balanced routes. 
Right: A multi-hazard NO$_2$ exposure example in Los Angeles, where satellite context and exposure artifacts support population-exposure estimation. The traces illustrate representative compressed agentic workflows.} 
    \label{fig:samples_tasks}
\end{figure}

\section{Related Work}

\paragraph{Disaster Geospatial Benchmarks.}
Early disaster remote-sensing benchmarks focused on static perception: damage 
labels in xBD/xView2~\cite{xbd,xview2}, flood masks in 
Sen1Floods11~\cite{sen1floods11}, post-flood segmentation in 
FloodNet~\cite{floodnet}, scene parsing in RescueNet~\cite{rescuenet}, and 
flooded-road extraction in SpaceNet~8~\cite{spacenet8}. Subsequent benchmarks 
expanded to multimodal question answering and geospatial interpretation, 
including FloodNet-VQA~\cite{floodnet}, VQA-Aid~\cite{vqaaid}, 
DisasterM3~\cite{disasterm3}, HRVQA~\cite{hrvqa}, RSVLM-QA~\cite{rsvlmqa}, 
and GeoMMBench~\cite{geommbbench}. However, these datasets evaluate final 
outputs only, missing procedural errors, tool-use validity, and execution 
coherence~\cite{core}. Recent work has therefore moved toward tool-augmented 
agentic evaluation: ThinkGeo~\cite{thinkgeo} for step-wise tool-augmented RS 
reasoning, OpenEarthAgent~\cite{openearthagent} for supervised geospatial 
reasoning traces, and GeoMMAgent~\cite{geommbbench} for multi-agent geoscience 
interpretation. RescueADI~\cite{rescueadi} introduced adaptive disaster 
interpretation, while GeoLLM-QA~\cite{geollmqa} studied UI-grounded geospatial 
interaction. GeoDisaster extends this direction to fully operational disaster 
analysis—covering evacuation routing, damage inventory, SAR/optical 
deforestation assessment, and multi-temporal flood evolution—where each 
instance includes structured tool-use trajectories, intermediate state 
summaries, and raster/vector artifacts, enabling evaluation of both final 
correctness and the validity of the underlying reasoning process.

\noindent \textbf{Tool-Augmented and Agentic Geospatial Reasoning.}
Recent RS agentic systems augment LLMs with domain-specific tools for scene 
classification and counting~\cite{rsagent}, change interpretation~\cite{
changeagent}, wildfire monitoring~\cite{wildfireagents}, and cross-modal 
observation analysis~\cite{earthagent}. ThinkGeo~\cite{thinkgeo} evaluates 
ReAct-style agents on structured RS tasks, while 
OpenEarthAgent~\cite{openearthagent} trains geospatial agents with supervised 
multi-step reasoning trajectories. Despite these advances, such systems rely on 
a monolithic agent to plan, select tools, execute analysis, and synthesize 
responses—limiting modularity and robustness on complex workflows. Multi-agent 
frameworks partially address this: GeoLLM-Squad~\cite{geollmsquad} separates 
orchestration from task execution via specialized sub-agents, and 
GeoMMAgent~\cite{geommbbench} combines retrieval, perception, and reasoning 
agents for geoscience interpretation. Our framework follows this direction but 
specializes it for operational disaster reasoning, supporting multi-turn 
structured execution over 18 disaster-specific tools with evaluation of both 
final answers and intermediate multi-agent trajectories.

\noindent \textbf{Trajectory Optimization and Multi-Agent Alignment.}
LLMs remain unreliable on long-horizon tasks involving tool use and 
intermediate artifacts~\cite{openearthagent}. Post-training approaches address 
this through supervised trajectory tuning—ToolLLM~\cite{toolllm}, 
AgentTuning~\cite{agenttuning}—and preference-based alignment via PPO~\cite{
ziegler2019fine,ouyang2022training}, DPO~\cite{rafailov2024dpo}, and 
GRPO~\cite{shao2024deepseekmath}. Recent extensions target trajectory-level 
optimization: GiGPO~\cite{gigpo} improves credit assignment through episode- 
and step-level grouping, SELAUR~\cite{selaur} introduces uncertainty-aware 
rewards for multi-step exploration, and GeoReason~\cite{georeason} enforces 
logical consistency between reasoning traces and outputs. In multi-agent 
settings, failures additionally arise from coordination breakdowns and 
inconsistent role execution~\cite{mast2025}. MAGRPO~\cite{llmcollabmarl}, 
M-GRPO~\cite{mgrpo}, and MHGPO~\cite{mhgpo} extend GRPO-style optimization to 
collaborative and hierarchical agent systems, while Dr.~MAS~\cite{drmas} 
stabilizes training through agent-wise normalization. Our Role-Contract 
Expectation Alignment builds on these foundations by combining failure-aware 
SFT with contract-conditioned reinforcement learning, explicitly supervising 
role-conditioned obligations at each step of orchestrated execution—a 
dimension not addressed by prior trajectory optimization methods.

\begin{table*}[t]
\centering
\footnotesize
\setlength{\tabcolsep}{4pt}
\renewcommand{\arraystretch}{1.15}
\caption{\textbf{GeoDisaster task families.} Each family is derived from 
public EO/GIS sources and targets a distinct stage of operational disaster 
geospatial reasoning, spanning five disaster families, with heterogeneous 
input modalities and structured answer targets.}
\label{tab:geodisaster_overview}
\vspace{4pt}
\resizebox{\textwidth}{!}{
\begin{tabular}{p{0.13\textwidth} 
                >{\centering\arraybackslash}p{0.03\textwidth} 
                p{0.27\textwidth} 
                p{0.27\textwidth} 
                p{0.22\textwidth}}
\toprule
\rowcolor{blue!10}
\textbf{Task Family} & \textbf{N} & \textbf{Task Nature} & 
\textbf{Inputs \& Artifacts} & \textbf{Sources \& Products} \\
\midrule

\textbf{Deforestation Monitoring}
& 384
& Forest-loss detection, area estimation, temporal change, false-positive 
  rejection, canopy assessment, optical--SAR concordance.
& Sentinel-2 optical, Sentinel-1 SAR VV/VH, Hansen tree-cover/loss masks, 
  optical/SAR time series, forest-change reports.
& Sentinel-1/2, Hansen GFC, GEE. \\

\rowcolor{teal!6}
\textbf{Multi-Hazard Analysis}
& 432
& Hazard measurement, severity labeling, cross-source agreement, exposure 
  estimation across wildfire, urban expansion, drought, air pollution.
& Raster hazard stacks, optical indices, land-cover, climate/atmospheric 
  layers, population and OSM layers, exposure summaries.
& MODIS, VIIRS/FIRMS, Landsat, Sentinel-2/5P, Dynamic World, GHSL, 
  WorldCover, CHIRPS, ERA5-Land, WorldPop, OSM. \\

\textbf{Building-Damage Assessment}
& 1005
& Damage inventory, severe-building extraction, region ranking, cross-scene 
  comparison, spatial concentration, scene-level reporting.
& Pre/post RGB imagery, building polygons, damage labels, masks, 
  $4{\times}4$ grids, damage summaries.
& xBD/xView2~\cite{xbd}. \\

\rowcolor{teal!6}
\textbf{Flood-Safe Routing}
& 600
& Reachable-shelter search, fastest/safest route comparison, flood-aware 
  routing, visual overlay rendering.
& Sentinel-2 RGB, flood extents, OSM road graphs, shelter and route 
  candidates, blocked-road layers, rendered overlays.
& Sentinel-2, Copernicus EMS, OSM, OSMnx~\cite{boeing2017osmnx}. \\

\textbf{SAR Flood Monitoring}
& 500
& Flood inventory, polygon geometry, spatial concentration, known-water 
  separation, diagnostic reporting.
& Sentinel-1 SAR chips, Sentinel-2 previews, JRC water references, flood 
  masks, polygon components, grid summaries.
& Sen1Floods11~\cite{sen1floods11}, Sentinel-1/2, JRC GSW. \\

\bottomrule
\end{tabular}}
\end{table*}

%GeoDisaster spans flood-aware evacuation routing, building-damage inventory, SAR/optical deforestation analysis, and multi-temporal flood evolution, requiring agents to combine satellite imagery, raster masks, vector geometry, disaster metadata, tool calls, and generated geospatial artifacts.  

\section{Dataset Construction}
\label{sec:dataset_construction}

GeoDisaster is a multi-task agentic benchmark for operational geospatial 
disaster reasoning, constructed from five public EO/GIS source families, each 
converted into standardized task instances with visible inputs, structured 
answer targets, and executable geospatial workflows.

\noindent\textbf{Source ingestion and standardization.}
We ingest and normalize public EO/GIS sources into standardized sample units 
per task family. \textit{Deforestation monitoring} combines Sentinel-2 optical, 
Sentinel-1 SAR, and Hansen GFC products~\cite{hansen2013high}. 
\textit{Multi-hazard analysis} integrates raster and exposure products across 
wildfire, urban expansion, drought, and air pollution, using Dynamic World and 
ESA WorldCover for land cover, WorldPop for population exposure, and OSM for 
infrastructure context. \textit{Building-damage assessment} uses xBD/xView2 
pre/post imagery with building polygons and damage labels~\cite{gupta2019xbd}. 
\textit{Flood-safe routing} combines Sentinel-2 with Copernicus EMS flood 
extents~\cite{copernicusEMS} and OSM road graphs~\cite{boeing2017osmnx}. 
\textit{SAR flood monitoring} uses Sentinel-1 chips, water references, and 
Sen1Floods11 labels~\cite{sen1floods11}. This unifies raster, vector, graph, 
image, and metadata evidence under a common format while preserving 
family-specific reasoning requirements (Table~\ref{tab:geodisaster_overview}).

\noindent\textbf{Quality control and derived evidence.}
Each sample is filtered for complete imagery, valid geometries, usable 
metadata, and sufficient spatial content. Retained samples yield 
family-specific supervision targets: forest-loss statistics and cross-modal 
agreement for deforestation; hazard summaries and exposure statistics for 
multi-hazard; damage inventories and grid rankings for building assessment; 
route-risk scores and map overlays for flood routing; and flood masks and 
diagnostic reports for SAR monitoring. Building footprint areas are computed 
from image-space polygons using scene ground-sampling distance, avoiding 
unreliable projection assumptions at chip level. GeoDisaster improves geographic coverage by combining curated AOIs from multiple public EO/GIS datasets spanning different hazard types, sensors, and regional contexts, while not claiming globally uniform coverage.

\begin{wraptable}{r}{0.48\columnwidth}
\vspace{-8pt}
\centering
\setlength{\abovecaptionskip}{2pt}
\setlength{\belowcaptionskip}{4pt}
\footnotesize
\setlength{\tabcolsep}{3pt}
\renewcommand{\arraystretch}{1.15}
\caption{\textbf{GeoDisaster statistics.}}
\label{tab:dataset_stats}
\vspace{2pt}
\begin{tabular}{lc}
\toprule
\rowcolor{blue!12}
\textbf{Statistic} & \textbf{Value} \\
\midrule
\rowcolor{blue!5}
Total instances          & 2,921       \\
Task families / Q-types  & 5 / 43      \\
\rowcolor{blue!5}
Train / Test split       & 2,629 / 292 \\
\midrule
\rowcolor{teal!10}
Avg.\ GT tool calls      & 5.00        \\
\rowcolor{teal!5}
Avg.\ conv.\ turns       & 29.10       \\
\rowcolor{teal!10}
Avg.\ plan steps         & 2.52        \\
\rowcolor{teal!5}
All-specialists trajectories & 42.07\%     \\
\rowcolor{teal!10}
Avg.\ artifact handles   & 3.51        \\
\rowcolor{teal!5}
Avg.\ path references    & 16.72       \\
\bottomrule
\end{tabular}
\vspace{-6pt}
\end{wraptable}
\noindent\textbf{Structured task and trajectory generation.}
For each sample, we construct a natural-language task, a structured answer 
schema, and a tool-grounded reasoning trajectory. Tasks require multi-step 
spatial reasoning executed through 18 disaster-oriented geospatial tools 
covering evidence access, spatial analysis, damage assessment, exposure 
estimation, and structured reporting. Reference answers are grounded in 
deterministic geospatial workflows and consistency checks, eliminating 
language-model annotation. Each instance is converted into an agentic 
trajectory comprising orchestrator planning, specialist-agent actions, tool 
calls, observations, intermediate state summaries, and final structured 
responses, supporting failure-aware SFT and execution-level evaluation. Figure~\ref{fig:samples_tasks} shows sample tasks and trajectories.

\noindent\textbf{Scale and complexity.} Table~\ref{tab:dataset_stats} summarizes key benchmark statistics. The 
trajectory complexity metrics reveal that GeoDisaster demands genuinely 
non-trivial multi-step reasoning: tasks require coordinated subgoal 
decomposition across specialist agents rather than repetitive single-tool 
queries, instances are grounded in rich multi-source EO/GIS evidence, and 
42.07\% of trajectories activate all three specialist agents—confirming that 
coordinated visual, geospatial, and planning reasoning is genuinely required 
rather than incidental. Full statistics are in the supplementary.

\section{Proposed Methodology}

We formulate disaster geospatial reasoning as an orchestrated multi-agent 
decision process. Given an input task $q$---comprising a natural-language query, 
remote-sensing observations, and geospatial context---the objective is to produce 
a grounded final response $y$ through intermediate planning, tool use, artifact 
generation, and state-aware synthesis. We define the agent set and trajectory as
\begin{equation} \mathcal{A}=\{A_i \mid i\in\{\mathrm{ORC},\mathrm{GA},\mathrm{VRA},\mathrm{PA}\}\}, \qquad \tau=\{(s_t,\beta_t,a_t,o_t,\gamma_t)\}_{t=1}^{T}. \label{eq:agent_trajectory} \end{equation}

where $s_t$ is the shared execution state, $\beta_t\in\mathcal{A}$ is the active 
agent at step $t$, $a_t$ is its action, $o_t$ is the resulting observation, and 
$\gamma_t$ is the execution contract governing the current decision. Each role-specific policy is conditioned on the current state and active role instructions, and the final response is 
generated from the completed trajectory as $y=f(\tau,q)$.

We propose a two-stage methodology for operational disaster geospatial 
reasoning. In the first stage, a contract-driven orchestrated multi-agent 
framework decomposes complex geospatial tasks across role-specialized agents, 
where every orchestrator--specialist interaction is governed by a typed 
execution contract $\gamma_t^i$ that specifies verifiable obligations at each 
step. In the second stage, Role-Contract Expectation Alignment (RCEA) optimizes 
agent behavior through failure-aware supervised fine-tuning followed by 
contract-grounded reinforcement learning over dense, step-level signals. The 
key insight is that execution contracts---unlike free-form delegation---produce 
\emph{observable, step-level violations} $v_t^i$ that enable role-conditioned 
credit assignment; a property that terminal-reward methods structurally cannot 
exploit. Together, these stages address the two principal failure modes of 
existing multi-agent RS systems: (i) unverifiable role execution leading to 
silent spatial errors, and (ii) misattributed learning signals that destabilize 
joint optimization. %(Figure~\ref{fig:method}).

\subsection{Contract-Driven Orchestrated Agentic Framework}

Our framework adopts a centralized orchestration design in which the 
orchestrator $A_{\mathrm{ORC}}$ governs the full execution lifecycle—task 
decomposition, agent routing, state tracking, recovery, and termination—while 
three role-specialized agents provide complementary capabilities. Think of 
$A_{\mathrm{ORC}}$ as a project manager that assigns subtasks to specialists 
but never executes them directly. $A_{\mathrm{GA}}$ executes geospatial 
operations (boundary retrieval, distance computation, map rendering); 
$A_{\mathrm{VRA}}$ handles image and raster reasoning (visual interpretation, 
object localization, segmentation, spectral index analysis); and $A_{\mathrm{PA}}$ 
supports numerical computation, filtering, planning logic, and response 
synthesis. Specialists do not communicate directly—all information exchange 
is mediated through $A_{\mathrm{ORC}}$ via a shared, mutable execution state 
$s_t$ that encodes task context, plan status, available artifacts, tool 
outputs, and execution history. 

\noindent\textbf{Execution contracts as the core design primitive.}
The central novelty is not multi-agent decomposition per se—which prior 
systems also employ~\cite{geollmsquad,geommbbench}—but the formalization of 
every orchestrator--specialist interaction as a \textit{typed execution 
contract}. In prior RS multi-agent systems, the orchestrator issues free-form 
instructions: a specialist receives a text description of its task and is free 
to produce any response. This gives specialists no verifiable obligations—an 
agent may produce a plausible-sounding but spatially invalid output with no 
mechanism for detection. Our contracts eliminate this ambiguity structurally. 
When $A_{\mathrm{ORC}}$ invokes specialist $A_i$ at step $t$, it issues a 
six-tuple
\begin{equation}
\gamma_t^i =
\bigl(
g_t^i,\;
\mathcal{I}_t^i,\;
\mathcal{Y}_t^i,\;
\mathcal{E}_t^{\mathrm{req},i},\;
\Sigma_t^i,\;
\kappa_t^i
\bigr),
\label{eq:contract}
\end{equation}
where $g_t^i$ is the assigned subgoal, $\mathcal{I}_t^i$ is the task context, $\mathcal{Y}_t^i \in {\footnotesize \{\texttt{field},\texttt{layer},\texttt{artifact},\texttt{text}\}}$ is the required output type, 
$\mathcal{E}_t^{\mathrm{req},i}$ specifies required evidence and artifact 
dependencies, $\Sigma_t^i$ is the output schema, and $\kappa_t^i$ defines 
completion and failure conditions. Crucially, $\gamma_t^i$ does not prescribe 
a unique tool—it specifies \emph{what} must be produced and \emph{under what 
constraints}, preserving specialist flexibility while making accountability 
formally checkable.

Given $\gamma_t^i$, the set of admissible actions for agent $A_i$ at step $t$ is
\begin{equation}
\mathcal{F}_t^i = \bigl\{a \in \mathcal{U}_i \;\big|\; 
(a,\,s_t) \models 
(\mathcal{Y}_t^i,\,\mathcal{E}_t^{\mathrm{req},i},\,\Sigma_t^i)
\bigr\},
\label{eq:admissible}
\end{equation}
where $\mathcal{U}_i$ is the full action space of agent $A_i$. A contract violation 
occurs whenever the specialist's output falls outside this admissible set:
\begin{equation}
v_t^i =
\mathbb{1}\bigl[a_t^i \notin \mathcal{F}_t^i\bigr],
\label{eq:violation}
\end{equation}
covering invalid action types, malformed arguments, missing evidence, schema 
violations, and premature completion. Violations are \emph{observable at every 
step}—the structural property that makes dense, role-conditioned alignment 
tractable, and the key advantage over terminal-reward methods that observe only the trajectory endpoint. 
The full execution pipeline is
\begin{equation}
\resizebox{0.95\linewidth}{!}{$
q \xrightarrow{A_{\mathrm{ORC}}}
\texttt{Plan} \to \texttt{CallAgent}(A_i) \to \texttt{Tool/Reasoning}
\to \texttt{StateUpdate} \to \cdots \to \texttt{Terminate}
$}
\label{eq:pipeline}
\end{equation}

\subsection{Failure-Aware Supervised Fine-Tuning}

Before reinforcement learning, each agent requires a stable execution prior. 
Directly imitating raw successful trajectories is insufficient: a trajectory 
may satisfy $R_{\mathrm{task}}(\tau)=1$ while containing local contract 
violations ($v_t^i=1$) that were compensated by downstream actions. Imitating 
such trajectories instills violation-tolerant behavior that degrades under 
distribution shift. We instead construct \emph{failure-corrected} supervision 
data: development rollouts are analyzed to identify the most frequent failure 
types—repeated plans and actions, invalid tool calls, stale state references, missing 
artifacts, contract violations, and premature termination—and each erroneous action is replaced, under the same state and contract, with a recovery action verifiable against the task schema, state validator, and reference artifacts. The corrected dataset for agent $A_i$ is
\begin{equation}
\mathcal{D}_i=\bigl\{(s_t,\,\gamma_t^i,\,a_t^{*}) \;\big|\; 
\beta_t=A_i,\; a_t^* \in \mathcal{F}_t^i\bigr\},
\label{eq:sft_dataset}
\end{equation}
where $a_t^{*} \in \mathcal{F}_t^i$ is the corrected target action guaranteed 
to satisfy the active contract (Eq.~\ref{eq:admissible}). The per-agent SFT 
objective is
\begin{equation}
\mathcal{L}_{\mathrm{SFT}}^{(i)}
=
-\,\mathbb{E}_{(s_t,\,\gamma_t^i,\,a_t^{*})\sim\mathcal{D}_i}
\!\left[\log \pi_{\theta_i}(a_t^{*}\mid s_t,\gamma_t^i)\right].
\label{eq:sft_loss}
\end{equation}
$A_{\mathrm{ORC}}$ is trained on planning, routing, and recovery; specialists 
are trained on valid tool use, artifact-aware reasoning, and schema-compliant 
outputs. This stage is necessary: without a competent execution prior, 
$\mathbb{E}[R_{\mathrm{task}}(\tau)] \approx 0$ under random initialization 
due to combinatorial action spaces and long horizons, making RL gradient 
estimation intractable. SFT alone, however, is imitation-bound and cannot 
optimize trajectory-level behavior or recover from states outside 
$\mathcal{D}_i$. We address this with RCEA.

\subsection{Role-Contract Expectation Alignment (RCEA)}

\paragraph{Why terminal rewards fail in multi-agent systems.}
Standard agentic alignment combines supervised imitation with terminal-outcome 
RL~\cite{schulman2017ppo,rafailov2024dpo,shao2024deepseekmath}. When a 
trajectory succeeds or fails, every agent that participated receives the same 
scalar reward—regardless of whether its individual actions were correct. This 
creates two structural failures.

\noindent\textbf{(i) Credit misattribution.} Under standard policy gradient, 
the gradient update for agent $A_i$ is
\begin{equation}
\nabla_{\theta_i} \mathcal{L} \propto 
R_{\mathrm{task}}(\tau)
\sum_{t:\,\beta_t=A_i}
\nabla_{\theta_i}\log\pi_{\theta_i}(a_t \mid s_t, \gamma_t^i),
\label{eq:terminal_gradient}
\end{equation}
where $R_{\mathrm{task}}(\tau) \in \{0,1\}$ is the binary terminal reward. 
This scales every action of $A_i$ by the same terminal scalar, regardless of 
whether individual actions satisfied $\gamma_t^i$. An agent that committed 
contract violations but was compensated by other agents still receives a 
positive gradient—a fundamentally incorrect credit signal.

\noindent\textbf{(ii) Reward scale heterogeneity.} Agents differ in 
activation frequency $f_i = |\{t : \beta_t = A_i\}| / T$, action space 
cardinality $|\mathcal{U}_i|$, and accumulated reward magnitude. A global baseline
$\bar{R} = \frac{1}{K}\sum_k R_{\mathrm{task}}(\tau_k)$ conflates reward
distributions across roles with different execution contexts, which can obscure
role-specific failures and produce poorly scaled advantage estimates~\cite{drmas}.

% causing high- and low-activation agents to 
% share the same normalization statistics despite receiving rewards from 
% different execution contexts.

Recent multi-agent GRPO extensions~\cite{mgrpo,mhgpo} reduce variance through 
grouped sampling but still operate on terminal outcomes and introduce neither 
per-step role obligations nor contract-level verification. RCEA resolves both 
failures by replacing the terminal scalar in Eq.~\ref{eq:terminal_gradient} 
with a \emph{dense, per-step, per-agent} contract satisfaction signal, and 
replacing global normalization with role-specific advantage statistics.

\subsubsection{Step-Level Contract Satisfaction Reward}

At each step $t$, contract $\gamma_t^i$ (Eq.~\ref{eq:contract}) induces a 
constraint set $\mathcal{C}_t^i = \{\phi_1,\ldots,\phi_{|\mathcal{C}_t^i|}\}$ 
for agent $A_i$, where each $\phi_j$ is a deterministic check over 
$(a_t^i,o_t,s_{t+1})$. The violation indicator and step-level satisfaction 
score are
\begin{equation}
z_{t,j}^{i}
=
\mathbb{1}\!\left[\phi_j(a_t^i,o_t,s_{t+1})=\texttt{False}\right],
\qquad
R_{\mathcal{C},t}^{i}
=
1 - \frac{1}{|\mathcal{C}_t^i|}
\sum_{j=1}^{|\mathcal{C}_t^i|}
z_{t,j}^{i}\;\in[0,1].
\label{eq:step_reward}
\end{equation}
$R_{\mathcal{C},t}^{i}=1$ iff all constraints are satisfied; 
$R_{\mathcal{C},t}^{i}=0$ iff all are violated. Since each $\phi_j$ is 
decidable at the step where the action is produced, this yields a \emph{dense, 
local} signal unavailable to terminal-reward methods. The trajectory-level 
contract reward, accumulated only over steps where $A_i$ was active, is
\begin{equation}
R_{\mathcal{C}}^i(\tau)
=
\sum_{t=1}^{T}
R_{\mathcal{C},t}^{i}\cdot\mathbb{1}[\beta_t = A_i].
\label{eq:contract_reward}
\end{equation}
This replaces the uniform terminal scalar in Eq.~\ref{eq:terminal_gradient} 
with a role-localized, step-decomposed signal: credit is strictly proportional 
to each agent's contract compliance at each step.

\subsubsection{Trajectory Efficiency Penalty}

A specialist could satisfy its local contract while contributing to redundant 
or unstable trajectories—for instance, by producing valid but uninformative 
tool calls that force unnecessary replanning. We therefore add a 
trajectory-level efficiency penalty. Let $L_\tau=|\tau|$ be the trajectory 
length and $N_{\mathrm{rep}}$, $N_{\mathrm{replan}}$, $N_{\mathrm{loop}}$ 
count repeated actions, unnecessary replanning steps, and recovery loops. The 
penalty and total role-conditioned reward for agent $A_i$ are
\begin{equation}
R_{\mathcal{E}}(\tau)
=
-\alpha_1 L_\tau
-\alpha_2(N_{\mathrm{rep}}+N_{\mathrm{replan}}+N_{\mathrm{loop}}),
\quad
R_i(\tau)
=
R_{\mathrm{task}}(\tau)
+\lambda_{\mathcal{C}} R_{\mathcal{C}}^i(\tau)
+\lambda_{\mathcal{E}} R_{\mathcal{E}}(\tau),
\label{eq:total_reward}
\end{equation}
with $\alpha_1,\alpha_2,\lambda_{\mathcal{C}},\lambda_{\mathcal{E}}>0$. The 
three components operate at different granularities and are complementary by 
construction: $R_{\mathrm{task}}$ supervises episode-level correctness; 
$R_{\mathcal{C}}^i$ supervises step-level role compliance; and 
$R_{\mathcal{E}}$ penalizes trajectory instability. No single component 
subsumes the others.

\noindent \textbf{Role-wise Advantage Normalization.}
To resolve reward scale heterogeneity, advantages are normalized 
role-specifically rather than globally. Given $K$ sampled trajectories 
$\{\tau_k\}_{k=1}^{K}$, the role-specific mean, standard deviation, and 
group-relative advantage are
\begin{equation}
\mu_i=\frac{1}{K}\!\sum_{k=1}^{K}R_i(\tau_k),
\quad
\varsigma_i=\sqrt{\frac{1}{K}\sum_{k=1}^{K}(R_i(\tau_k)-\mu_i)^2},
\quad
\hat{A}_i(\tau_k)=\frac{R_i(\tau_k)-\mu_i}{\varsigma_i+\epsilon},
\label{eq:advantage}
\end{equation}
where $\epsilon>0$ is a small stability constant. The per-agent policy 
objective uses a clipped importance-ratio with KL regularization. For agent 
$A_i$, let $r_t^i = \pi_{\theta_i}(a_t \mid s_t,\gamma_t^i)\,/\,
\pi_{\mathrm{old}}(a_t \mid s_t,\gamma_t^i)$ be the importance ratio between 
the current and reference policy. The objective is

\begin{equation}
\mathcal{L}_{\mathrm{RL}}^{(i)}
=
-
\mathbb{E}_{\tau_k}
\!\left[
\sum_{t:\,\beta_t=A_i}
\min\!\left(
r_t^i\,\hat{A}_i(\tau_k),\;
\mathrm{clip}(r_t^i,\,1{-}\delta,\,1{+}\delta)\,
\hat{A}_i(\tau_k)
\right)
\right]
+
\beta_{\mathrm{KL}}\,
D_{\mathrm{KL}}\!\left(
\pi_{\theta_i}\,\|\,\pi_{\mathrm{ref}}
\right).
\label{eq:rl_objective}
\end{equation}

where $\delta>0$ is the clipping threshold and $\beta_{\mathrm{KL}}>0$ 
controls the deviation from the reference policy $\pi_{\mathrm{ref}}$. Each 
policy is updated exclusively on the actions it produced, with advantages 
normalized within its own role's reward distribution. This preserves role-specific failure signals in the advantage estimates and yields appropriately scaled gradients for stable joint optimization across heterogeneous agents.

\section{Experimental Evaluations}

\noindent\textbf{Benchmarks.}
We evaluate on two agentic RS benchmarks. \textbf{GeoDisaster} assesses 
disaster-oriented geospatial reasoning under multi-step tool use, constraint 
satisfaction, and operational decision-making. \textbf{OpenEarthAgent}~\cite{
openearthagent} evaluates generalization on an existing tool-augmented RS 
benchmark with broad geospatial task coverage. %Since ThinkGeo~\cite{thinkgeo} is subsumed by OpenEarthAgent, we do not report it separately.

\noindent\textbf{Baselines.}
We compare closed- and open-source LLMs, the OpenEarthAgent agentic baseline, where applicable, 
and three variants of our framework. Single-model baselines use structured 
prompting with tool descriptions but without role decomposition or delegation. 
Our variants are: \textbf{\textit{MAS}}—vanilla multi-agent system without alignment; 
\textbf{\textit{MAS+SFT}}—with failure-aware supervised fine-tuning applied per agent; 
and \textbf{\textit{MAS+SFT+GRPO}}—the full proposed framework with RCEA.

\noindent\textbf{Fairness Protocol.}
All methods share the same train/validation/test split, tool registry, tool 
documentation, execution budget, maximum tool calls, prompts, and stopping 
criteria. We use \textit{Qwen2.5-7B-Inst} as the default agent backbone. 
Both SFT and GRPO are trained on GeoDisaster training instances plus 
a balanced subset of OpenEarthAgent training data.

\noindent\textbf{Evaluation Metrics.}
We report both end-to-end and step-wise metrics to evaluate full trajectory 
quality beyond final-answer correctness. End-to-end metrics cover tool-sequence 
fidelity under order-agnostic, exact-order, and unique-tool matching 
(\textit{ToolAnyOr}, \textit{ToolSameO}, \textit{ToolUni}), task success 
(\textit{TSR}), answer correctness (\textit{Ans}), generative output quality 
(\textit{Gen}), contract satisfaction (\textit{CSR}), and strict episode 
completion (\textit{ESR}). Step-wise metrics cover instruction following 
(\textit{Inst.}), tool selection (\textit{Tool}), argument-name completeness 
(\textit{ArgN}), argument-value correctness (\textit{ArgV}), and execution 
summarization (\textit{Summa}). All metrics are scored using GPT-5.5 as an 
LLM-as-judge; Deeper description provided in the supplementary.

\subsection{Main Results}

Table~\ref{tab:geodisaster_results} reveals a stark capability gap between 
standalone LLMs and our aligned framework on \textbf{GeoDisaster}. Open-source LLMs fail almost 
entirely—near-zero tool-sequence fidelity and answer accuracy confirm that 
multi-step geospatial reasoning, structured tool invocation, and constraint 
satisfaction cannot emerge from prompting alone. Closed-source models perform 
better: \textit{GPT-5.5} achieves the strongest single-model answer accuracy 
(61.99) but still exhibits low tool-chain fidelity (ToolAnyOr: 20.93), 
indicating that correct final answers often arise despite—not because 
of—valid tool execution. \textit{o4-mini} shows the inverse: reasonable tool 
coverage but poor answer accuracy, suggesting aggressive but imprecise tool 
use.

Among our variants, \textit{MAS} already surpasses all open-source baselines 
without any alignment, confirming that role-based orchestration improves tool 
selection and argument construction over monolithic prompting. However, low 
Summa (30.03) and CSR (18.26) expose that decomposition alone cannot enforce 
contract compliance or grounded evidence synthesis. \textit{MAS+SFT} delivers 
the largest single gain—pushing tool fidelity above 98 and answer accuracy to 
82.43—by instilling role-specific execution priors and contract-compliant 
action patterns. \textit{MAS+SFT+GRPO} further refines trajectory-level 
behavior: answer accuracy improves to \textbf{90.11}, TSR to \textbf{94.24}, 
and step-wise metrics reach \textbf{100.00} across Inst., Tool, ArgN, and 
ArgV—demonstrating that dense, contract-grounded RL substantially reduces residual 
execution failures that imitation alone cannot resolve.

\begin{table*}[t]
\centering
\scriptsize
\setlength{\tabcolsep}{3.5pt}
\renewcommand{\arraystretch}{1.12}
\caption{\textbf{GeoDisaster benchmark results.} Tool-use fidelity: 
order-agnostic (\textit{ToolAnyOr}), exact-order (\textit{ToolSameO}), and 
unique-tool (\textit{ToolUni}) matching. \textit{TSR}: task success rate; 
\textit{Ans}: answer accuracy; \textit{CSR}: contract satisfaction; 
\textit{ESR}: strict episode resolution. Step-wise: instruction following 
(\textit{Inst.}), tool selection (\textit{Tool}), argument names 
(\textit{ArgN}), argument values (\textit{ArgV}), summarization 
(\textit{Summa}). Best results in \textbf{bold}.}
\label{tab:geodisaster_results}
\vspace{4pt}
\resizebox{\textwidth}{!}{
\begin{tabular}{l|c|cccccc|cccccc}
\toprule
\rowcolor{blue!10}
\textbf{Model} & \textbf{Type}
& \multicolumn{6}{c|}{\textbf{End-to-end Metrics}}
& \multicolumn{6}{c}{\textbf{Step-wise Metrics}} \\
\cmidrule(lr){3-8}\cmidrule(lr){9-14}
\rowcolor{blue!5}
& & \textit{ToolAnyOr} & \textit{ToolSameO} & \textit{ToolUni} 
& \textit{TSR} & \textit{Ans} & \textit{CSR}
& \textit{ESR} & \textit{Inst.} & \textit{Tool} 
& \textit{ArgN} & \textit{ArgV} & \textit{Summa} \\
\midrule

\rowcolor{gray!8}
\multicolumn{14}{l}{\textbf{Closed-source LLMs}} \\
GPT-5.5        & Single & 20.93 & 20.93 & 30.23 & 55.06 & 61.99 & 71.44
                        & 100.00 & 100.00 & 95.86 & 95.86 & 95.86 & 82.32 \\
\rowcolor{blue!3}
GPT-5          & Single & 11.99 & 11.99 & 13.70 & 46.27 & 60.48 & 65.37
                        & 81.51  & 81.51  & 69.71 & 71.56 & 69.71 & 69.27 \\
GPT-4o         & Single &  3.44 &  3.44 &  3.78 & 42.65 & 60.19 & 65.43
                        & 80.76  & 80.76  & 78.53 & 78.53 & 78.53 & 71.23 \\
\rowcolor{blue!3}
o4-mini        & Single & 23.26 & 20.93 & 23.26 & 29.12 & 26.64 & 30.59
                        & 44.19  & 44.19  & 86.97 & 86.97 & 86.97 & 47.10 \\

\midrule
\rowcolor{gray!8}
\multicolumn{14}{l}{\textbf{Open-source LLMs}} \\
Qwen2.5-7B-Inst.  & Single &  2.40 & 0.34 & 4.45 & 3.11 & 2.53 & 4.45
                           &  5.82 &  5.82 & 42.06 & 77.05 & 36.52 &  6.45 \\
\rowcolor{blue!3}
Qwen3-4B-Inst.    & Single &  0.00 & 0.00 & 0.00 & 1.16 & 1.16 & 2.33
                           &  2.33 &  6.98 & 40.00 & 23.26 &  9.30 &  3.72 \\
Llama-3.1-8B-Inst.& Single &  0.00 & 0.00 & 0.00 & 0.15 & 0.25 & 0.25
                           &  0.25 &  0.25 &  0.00 &  0.70 &  0.35 &  0.35 \\
\rowcolor{blue!3}
Mistral-7B-Inst.  & Single &  0.00 & 0.00 & 0.00 & 0.15 & 0.25 & 0.25
                           &  0.25 &  0.25 &  0.00 &  0.70 &  0.35 &  0.35 \\
Qwen2.5-VL-7B     & Single &  0.00 & 0.00 & 0.00 & 0.95 & 1.50 & 2.10
                           &  1.16 &  3.49 &  0.00 & 12.40 &  5.81 &  3.10 \\

\midrule
\rowcolor{gray!8}
\multicolumn{14}{l}{\textbf{Ours}} \\
MAS            & Multi  &  6.85 &  4.11 & 10.62 & 20.36 & 22.93 & 18.26
                        & 49.66  & 90.88  & 81.71 & 95.46 & 62.57 & 30.03 \\
\rowcolor{teal!6}
MAS+SFT        & Multi  & 98.75 & 98.26 & 98.41 & 92.46 & 82.43 & 88.16
                        & 94.69  & 99.43  & 99.58 & 99.58 & 99.58 & 91.01 \\
\rowcolor{teal!12}
MAS+SFT+GRPO   & Multi  
& \textbf{98.56} & \textbf{97.76} & \textbf{98.72} 
& \textbf{94.24} & \textbf{90.11} & \textbf{88.88}
& \textbf{95.89} & \textbf{100.00} & \textbf{100.00} 
& \textbf{100.00} & \textbf{100.00} & \textbf{91.59} \\
\bottomrule
\end{tabular}}
\end{table*}

\noindent Table~\ref{tab:openearthagent_results} evaluates generalization and gain transferability of our framework on 
\textbf{OpenEarthAgent}; baseline results are taken from the original paper as the 
benchmark agent is not publicly available. Open-source LLMs perform relatively 
better here than on GeoDisaster, corroborating that GeoDisaster imposes 
stricter demands on tool-call construction and constraint satisfaction. Single 
models nonetheless remain limited: even the strongest closed-source model 
(GPT-4o) reaches only 45.80 ArgV, exposing a persistent gap between 
instruction-following and correct argument execution.

Our aligned framework outperforms both single-model LLMs and the 
OpenEarthAgent agentic baseline across all primary metrics. Against 
OpenEarthAgent, \textit{MAS+SFT+GRPO} improves answer accuracy by 
$+$\textbf{39.9} points (45.26$\to$85.16), generative output by 
$+$\textbf{18.5} points (75.86$\to$94.39), unique-tool fidelity by 
$+$\textbf{20.6} points (72.71$\to$93.33), and argument-value correctness by 
$+$\textbf{33.4} points (62.10$\to$95.45). \textit{MAS+SFT} drives the 
primary gains in tool use and argument construction; \textit{GRPO} 
subsequently reduces residual loops, replanning errors, and execution 
instability. Crucially, ArgN reaches \textbf{100.00} after SFT and is 
maintained by GRPO, indicating that contract-conditioned fine-tuning was able to fully 
resolve argument-name errors. These results confirm that role-specialized 
orchestration and RCEA transfer effectively beyond GeoDisaster to a 
structurally different RS agentic benchmark.

\begin{table*}[t]
\centering
\scriptsize
\setlength{\tabcolsep}{3.5pt}
\renewcommand{\arraystretch}{1.12}
\caption{\textbf{OpenEarthAgent benchmark results.} Tool-use fidelity: 
order-agnostic (\textit{ToolAnyOr}), exact-order (\textit{ToolSameO}), 
unique-tool (\textit{ToolUni}) matching. \textit{Ans}: non-generative 
accuracy; \textit{Gen}: generative accuracy. Step-wise: instruction following 
(\textit{Inst.}), tool selection (\textit{Tool}), argument names 
(\textit{ArgN}), argument values (\textit{ArgV}), summarization 
(\textit{Summa}).}
\label{tab:openearthagent_results}
\vspace{4pt}
\resizebox{\textwidth}{!}{
\begin{tabular}{l|c|ccccc|ccccc}
\toprule
\rowcolor{blue!10}
\textbf{Model} & \textbf{Type}
& \multicolumn{5}{c|}{\textbf{End-to-end Metrics}}
& \multicolumn{5}{c}{\textbf{Step-wise Metrics}} \\
\cmidrule(lr){3-7}\cmidrule(lr){8-12}
\rowcolor{blue!5}
& & \textit{ToolAnyOr} & \textit{ToolSameO} & \textit{ToolUni} 
& \textit{Ans} & \textit{Gen}
& \textit{Inst.} & \textit{Tool} & \textit{ArgN} 
& \textit{ArgV} & \textit{Summa} \\
\midrule

\rowcolor{gray!8}
\multicolumn{12}{l}{\textbf{Closed-source LLMs}} \\
GPT-5          & Single & 46.96 & 46.79 & 47.81 & 43.88 & 46.21
                        & 97.54 & 82.69 & 73.28 & 37.28 & 87.02 \\
\rowcolor{blue!3}
GPT-4o         & Single & 50.81 & 50.38 & 55.52 & 39.22 & 77.93
                        & 99.18 & 93.88 & 85.48 & 45.80 & 86.76 \\
o4-mini        & Single & 40.12 & 39.95 & 41.49 & 35.18 & 55.17
                        & 83.68 & 68.33 & 64.17 & 37.95 & \textbf{89.48} \\

\midrule
\rowcolor{gray!8}
\multicolumn{12}{l}{\textbf{Open-source LLMs}} \\
Qwen2.5-7B-Inst.   & Single & 31.57 & 30.02 & 36.61 & 15.85 & 41.38
                            & 94.08 & 85.51 & 78.46 & 38.00 & 80.08 \\
\rowcolor{blue!3}
Qwen3-4B-Inst.     & Single & 16.00 & 14.71 & 21.47 & 13.72 & 15.86
                            & 97.34 & 86.94 & 84.12 & 33.55 & 83.28 \\
Qwen2.5-3B-Inst.   & Single &  9.24 &  8.72 & 12.40 &  1.83 & 24.14
                            & 85.07 & 72.13 & 64.87 & 24.12 & 68.75 \\
\rowcolor{blue!3}
Llama-3.1-8B-Inst. & Single & 39.01 & 37.72 & 44.91 & 12.70 & 55.17
                            & 47.07 & 39.30 & 34.11 & 17.49 & 79.08 \\
InternLM3-8B-Inst. & Single &  2.22 &  1.88 &  3.16 &  0.24 &  2.76
                            & 44.54 & 38.46 & 27.82 & 13.25 & 29.84 \\
\rowcolor{blue!3}
Mistral-7B-Inst.   & Single &  2.65 &  2.57 &  2.91 &  0.39 &  5.52
                            & 64.14 & 35.12 & 26.11 & 12.45 & 24.04 \\

\midrule
\rowcolor{gray!8}
\multicolumn{12}{l}{\textbf{Agentic Baseline}} \\
OpenEarthAgent     & Multi  & 67.75 & 67.24 & 72.71 & 45.26 & 75.86
                            & 99.51 & \textbf{97.18} & 96.08 & 62.10 & 83.64 \\

\midrule
\rowcolor{gray!8}
\multicolumn{12}{l}{\textbf{Ours}} \\
MAS+SFT            & Multi  & 73.89 & 65.56 & 85.00 & 80.24 & 85.46
                            & 99.57 & 96.92 & \textbf{100.00} & 94.70 & 81.17 \\
\rowcolor{teal!12}
MAS+SFT+GRPO       & Multi  
& \textbf{80.56} & \textbf{75.56} & \textbf{93.33} 
& \textbf{85.16} & \textbf{94.39}
& \textbf{99.80} & 96.76 & \textbf{100.00} 
& \textbf{95.45} & 86.80 \\
\bottomrule
\end{tabular}}
\end{table*}

\subsection{Further Analysis}

Beyond aggregate results, we conduct four targeted analyses to understand 
\textit{where} failures originate, \textit{which} agents benefit most from 
alignment, \textit{how} robust the framework is across backbones, and 
\textit{what} the computational cost of reliability is.

\noindent\textbf{Agent-Specific Execution Analysis.}
Table~\ref{tab:agent_metrics} breaks down step-wise metrics by agent role.
\textit{MAS} establishes valid execution structure—high ArgN across all roles 
confirms that orchestration correctly scaffolds argument naming—but exposes 
critical gaps in tool selection (ORC: 57.49, VRA: 84.72) and argument-value 
correctness (ORC: 62.14, GA: 72.75), showing that role decomposition alone 
cannot enforce semantic correctness. \textit{MAS+SFT} pushes most metrics to 
saturation: VRA and PA reach 100.00 across all step-wise metrics, confirming 
that failure-corrected supervision teaches contract-compliant execution 
per role. The main residual is summarization quality (ORC: 91.01, PA: 91.58), 
reflecting the difficulty of grounded synthesis beyond local action 
correctness. \textit{MAS+SFT+GRPO} eliminates remaining tool and argument 
errors in ORC and GA, reaching 100.00 for VRA and GA on all metrics—with 
summarization as the sole persistent gap and primary target for future work.

\begin{table*}[t]
\centering
\scriptsize
\setlength{\tabcolsep}{3.5pt}
\renewcommand{\arraystretch}{1.12}
\caption{\textbf{Agent-specific step-wise metrics on GeoDisaster.} ORC: orchestrator; 
VRA: visual reasoning; GA: geospatial analysis; PA: planning. 
\textit{Inst}: instruction following; \textit{Tool}: tool selection; 
\textit{ArgN/ArgV}: argument name/value correctness; 
\textit{Summa}: summarization.}
\label{tab:agent_metrics}
\vspace{4pt}
\resizebox{\textwidth}{!}{
\begin{tabular}{l|ccccc|ccccc|ccccc|ccccc}
\toprule
\rowcolor{blue!10}
\textbf{Model}
& \multicolumn{5}{c|}{\textbf{ORC}}
& \multicolumn{5}{c|}{\textbf{VRA}}
& \multicolumn{5}{c|}{\textbf{GA}}
& \multicolumn{5}{c}{\textbf{PA}} \\
\cmidrule(lr){2-6}\cmidrule(lr){7-11}
\cmidrule(lr){12-16}\cmidrule(lr){17-21}
\rowcolor{blue!5}
& \textit{Inst} & \textit{Tool} & \textit{ArgN} 
& \textit{ArgV} & \textit{Summa}
& \textit{Inst} & \textit{Tool} & \textit{ArgN} 
& \textit{ArgV} & \textit{Summa}
& \textit{Inst} & \textit{Tool} & \textit{ArgN} 
& \textit{ArgV} & \textit{Summa}
& \textit{Inst} & \textit{Tool} & \textit{ArgN} 
& \textit{ArgV} & \textit{Summa} \\
\midrule
\rowcolor{gray!5}
MAS
& 94.85 & 57.49 & 94.85 & 62.14 & 30.03
& 99.66 & 84.72 & 99.50 & 67.36 & 66.37
& 99.50 & 96.11 & 99.50 & 72.75 & 30.02
& 99.90 & 88.51 & 99.50 & 78.97 & 60.98 \\
\rowcolor{teal!6}
MAS+SFT
& 99.77 & 97.24 & 99.77 & 97.46 & 91.01
& 100.00 & 100.00 & 100.00 & 100.00 & 100.00
& 99.07 & 99.32 & 99.32 & 99.32 & 99.07
& 100.00 & 100.00 & 100.00 & 100.00 & 91.58 \\
\rowcolor{teal!14}
MAS+SFT+GRPO
& \textbf{100.00} & \textbf{97.95} & \textbf{100.00} 
& \textbf{97.95} & \textbf{91.59}
& \textbf{100.00} & \textbf{100.00} & \textbf{100.00} 
& \textbf{100.00} & \textbf{100.00}
& \textbf{100.00} & \textbf{100.00} & \textbf{100.00} 
& \textbf{100.00} & \textbf{100.00}
& \textbf{100.00} & \textbf{100.00} & \textbf{100.00} 
& \textbf{100.00} & \textbf{91.59} \\
\bottomrule
\end{tabular}}
\end{table*}

\noindent\textbf{Backbone Sensitivity Analysis.}
Table~\ref{tab:backbone_analysis} evaluates RCEA across four LLM backbones.
\textbf{Key finding:} orchestration-level metrics (PlanAcc, DelegAcc) saturate 
above 97 for all backbones, confirming the protocol is backbone-agnostic for 
subgoal assignment. Performance diverges on metrics requiring grounded 
reasoning: \textit{Qwen2.5-7B} achieves the best balance (Ans: \textbf{90.11}, 
TSR: \textbf{94.24}, Summa: \textbf{91.59}); \textit{Qwen3-4B} remains 
competitive despite its smaller size (Ans: 77.28, near-saturated Tool and 
ArgN), confirming alignment effectiveness on lighter backbones. 
\textit{Mistral} attains the highest PlanAcc (99.44) and perfect ToolUni 
(100.00) but lags in Ans (73.97) and Summa (78.96)—a dissociation between 
structural execution and grounded synthesis. \textit{Llama} shows the largest 
tool-chain fidelity drop (ToolAnyOr: 76.74). \textbf{Takeaway:} framework 
design is robust across backbones; backbone reasoning capacity is the binding 
constraint for answer accuracy and summarization.

\begin{table*}[t]
\centering
\scriptsize
\setlength{\tabcolsep}{3.5pt}
\renewcommand{\arraystretch}{1.12}
\caption{\textbf{Backbone sensitivity on GeoDisaster (MAS+SFT+GRPO).} 
\textit{PlanAcc/DelegAcc}: planning and delegation accuracy; 
\textit{ToolAnyOr/SameO/Uni}: tool-use fidelity; \textit{TSR}: task success; 
\textit{Ans}: answer accuracy; \textit{CSR}: constraint satisfaction. 
Step-wise as in Table~\ref{tab:agent_metrics}.}
\label{tab:backbone_analysis}
\vspace{4pt}
\resizebox{\textwidth}{!}{
\begin{tabular}{l|cccccccc|ccccc}
\toprule
\rowcolor{blue!10}
\textbf{Backbone}
& \multicolumn{8}{c|}{\textbf{End-to-end Metrics}}
& \multicolumn{5}{c}{\textbf{Step-wise Metrics}} \\
\cmidrule(lr){2-9}\cmidrule(lr){10-14}
\rowcolor{blue!5}
& \textit{PlanAcc} & \textit{DelegAcc}
& \textit{ToolAnyOr} & \textit{ToolSameO} & \textit{ToolUni}
& \textit{TSR} & \textit{Ans} & \textit{CSR}
& \textit{Inst.} & \textit{Tool} & \textit{ArgN} 
& \textit{ArgV} & \textit{Summa} \\
\midrule
\rowcolor{teal!12}
Qwen2.5-7B-Instruct
& 97.95 & 97.95
& \textbf{98.56} & \textbf{97.76} & 98.72
& \textbf{94.24} & \textbf{90.11} & \textbf{88.88}
& \textbf{100.00} & \textbf{100.00} & \textbf{100.00} 
& \textbf{100.00} & \textbf{91.59} \\
\rowcolor{teal!5}
Qwen3-4B-Instruct
& 97.93 & 97.92
& 95.89 & 95.89 & 95.89
& 90.34 & 77.28 & 86.76
& 99.62 & 99.40 & 99.78 & 99.57 & 90.03 \\
Llama-3.1-8B-Instruct
& 98.37 & 97.11
& 76.74 & 76.74 & 81.40
& 77.22 & 67.41 & 71.00
& 99.54 & 96.86 & 99.94 & 99.89 & 72.40 \\
\rowcolor{blue!3}
Mistral-7B-Instruct-v0.3
& \textbf{99.44} & \textbf{99.44}
& 90.70 & 90.70 & \textbf{100.00}
& 85.90 & 73.97 & 77.56
& 99.33 & 99.98 & 99.94 & 99.89 & 78.96 \\
\bottomrule
\end{tabular}}
\end{table*}

\noindent\textbf{Failure-Wise Analysis.}
Table~\ref{tab:failure_analysis} reports per-category error rates (lower is 
better). Open-source LLMs fail catastrophically—ToolErr, FormatErr, TermErr, 
and ConstraintErr all exceed 90—confirming that prompting alone cannot sustain 
structured geospatial workflows. Closed-source models eliminate formatting and 
loop errors but maintain high ToolErr (GPT-4o: 96.56, GPT-5: 88.01), showing 
persistent failure to recover the required tool chain regardless of model 
scale. \textit{MAS} substantially reduces FormatErr but amplifies coordination 
failures: PlanErr (39.33), AgentErr (37.86), LoopErr (36.99), and ReplanErr 
(77.74) all increase—confirming that decomposition without alignment worsens 
coordination. \textit{MAS+SFT} surgically resolves the highest-impact 
categories: PlanErr (39.33$\to$2.76), ToolErr (93.15$\to$1.37), ArgErr 
(37.43$\to$0.00), FormatErr and LoopErr reach zero. \textit{MAS+SFT+GRPO} 
eliminates residual ReplanErr, ToolExecErr, and AbortErr via trajectory-level optimization. 

% \textbf{Persistent residuals:} SynthErr (8.41) and 
% ConstraintErr (11.12) remain, requiring grounded synthesis and explicit 
% constraint verification beyond local action correction.

\begin{table*}[t]
\centering
\scriptsize
\setlength{\tabcolsep}{3pt}
\renewcommand{\arraystretch}{1.12}
\caption{\textbf{Failure-wise error rates on GeoDisaster} (lower is better). 
Each cell: \% of trajectories where the error materially affected execution. 
\textit{PlanErr}: planning; \textit{AgentErr}: agent selection; 
\textit{ToolErr}: tool selection; \textit{ArgErr}: arguments; 
\textit{FormatErr}: output format; \textit{TermErr}: termination; 
\textit{LoopErr}: repeated calls; \textit{ReplanErr}: replanning; 
\textit{SynthErr}: synthesis; \textit{ConstraintErr}: constraint violation; 
\textit{ToolExecErr}: tool execution; \textit{AbortErr}: aborted trajectory.}
\label{tab:failure_analysis}
\vspace{4pt}
\resizebox{\textwidth}{!}{
\begin{tabular}{l|c|cccccccccccc}
\toprule
\rowcolor{blue!10}
\textbf{Model} & \textbf{Type}
& \textit{PlanErr} & \textit{AgentErr} & \textit{ToolErr} & \textit{ArgErr}
& \textit{FormatErr} & \textit{TermErr} & \textit{LoopErr} & \textit{ReplanErr}
& \textit{SynthErr} & \textit{ConstraintErr} & \textit{ToolExecErr} 
& \textit{AbortErr} \\
\midrule
\rowcolor{gray!8}
\multicolumn{14}{l}{\textbf{Closed-source LLMs}} \\
GPT-5.5 & Single
& -- & -- & 79.07 & 4.14 & \textbf{0.00} & \textbf{0.00} 
& \textbf{0.00} & \textbf{0.00} & 17.68 & 28.56 & 4.14 & \textbf{0.00} \\
\rowcolor{blue!3}
GPT-5 & Single
& -- & -- & 88.01 & 30.29 & \textbf{0.00} & 18.49 
& 18.49 & 18.49 & 30.73 & 34.63 & 30.29 & 18.49 \\
GPT-4o & Single
& -- & -- & 96.56 & 21.47 & 13.40 & 19.24 
& 5.84 & 5.84 & 28.77 & 34.57 & 21.47 & 19.24 \\
\rowcolor{blue!3}
o4-mini & Single
& -- & -- & 76.74 & 13.03 & 25.58 & 55.81 
& 30.23 & 30.23 & 52.90 & 69.41 & 13.03 & 55.81 \\
\midrule
\rowcolor{gray!8}
\multicolumn{14}{l}{\textbf{Open-source LLMs}} \\
Qwen2.5-7B-Inst. & Single
& -- & -- & 97.60 & 63.48 & 94.18 & 94.18 
& \textbf{0.00} & \textbf{0.00} & 93.55 & 95.55 & 57.94 & 94.18 \\
\rowcolor{blue!3}
Qwen3-4B-Inst. & Single
& -- & -- & 100.00 & 90.70 & 93.02 & 97.67 
& \textbf{0.00} & \textbf{0.00} & 96.28 & 97.67 & 60.00 & 97.67 \\
Llama-3.1-8B-Inst. & Single
& -- & -- & 100.00 & 99.65 & 99.75 & 99.75 
& \textbf{0.00} & \textbf{0.00} & 99.65 & 99.75 & 100.00 & 99.75 \\
\rowcolor{blue!3}
Mistral-7B-Inst. & Single
& -- & -- & 100.00 & 99.65 & 99.75 & 99.75 
& \textbf{0.00} & \textbf{0.00} & 99.65 & 99.75 & 100.00 & 99.75 \\
Qwen2.5-VL-7B & Single
& -- & -- & 100.00 & 94.19 & 96.51 & 98.84 
& \textbf{0.00} & \textbf{0.00} & 96.90 & 97.90 & 100.00 & 98.84 \\
\midrule
\rowcolor{gray!8}
\multicolumn{14}{l}{\textbf{Ours}} \\
MAS & Multi
& 39.33 & 37.86 & 93.15 & 37.43 & 9.12 & 50.34 
& 36.99 & 77.74 & 69.97 & 81.74 & 17.50 & 50.34 \\
\rowcolor{teal!6}
MAS+SFT & Multi
& 2.76 & 2.54 & 1.37 & \textbf{0.00} & \textbf{0.00} & 5.31 
& \textbf{0.00} & 0.34 & 8.99 & 11.84 & 1.37 & 1.03 \\
\rowcolor{teal!14}
MAS+SFT+GRPO & Multi
& \textbf{2.05} & \textbf{2.05} & \textbf{1.24} & \textbf{0.00} 
& \textbf{0.00} & \textbf{4.11} & \textbf{0.00} & \textbf{0.00} 
& \textbf{8.41} & \textbf{11.12} & \textbf{0.00} & \textbf{0.00} \\
\midrule
\rowcolor{gray!8}
\multicolumn{14}{l}{\textbf{Ours with Different Backbones (SFT+GRPO)}} \\
\rowcolor{blue!3}
Qwen3-4B-Inst. & Multi
& 2.07 & 2.08 & 4.11 & 0.43 & 0.38 & 4.14 
& \textbf{0.00} & \textbf{0.00} & 9.97 & 13.24 & 0.60 & 4.14 \\
Llama-3.1-8B-Inst. & Multi
& 1.63 & 2.89 & 23.26 & 0.11 & 0.46 & 5.77 
& \textbf{0.00} & 6.98 & 27.60 & 29.00 & 3.14 & 5.77 \\
\rowcolor{blue!3}
Mistral-7B-Inst. & Multi
& 0.56 & 0.56 & 9.30 & 0.11 & 0.67 & 3.45 
& \textbf{0.00} & \textbf{0.00} & 21.04 & 22.44 & 0.02 & 3.45 \\
\bottomrule
\end{tabular}}
\end{table*}

\noindent\textbf{Efficiency Analysis.}
Table~\ref{tab:efficiency_analysis} examines the accuracy--cost trade-off.
Near-zero Ans with zero Avg. Tools in weak models reflects execution collapse, 
not efficiency. Among high-performing variants, \textit{MAS+SFT} vs 
\textit{MAS+SFT+GRPO} is the critical comparison: despite identical Avg. 
Agents (3.42) and near-identical tool and token usage ($\sim$28K tokens), 
GRPO reduces Tokens/Succ. by 8.4\% (34,790$\to$31,859), Latency/Succ. by 
46.8\% (57.64$\to$30.67), and Calls/Succ. by 8.9\% (5.96$\to$5.43) while 
improving Ans by $+$7.7 points. \textbf{Takeaway:} GRPO improves trajectory 
discipline so more executions succeed, reducing success-normalized cost rather 
than adding overhead. Across SFT+GRPO backbones, Qwen2.5-7B achieves the best 
Latency/Succ. (30.67) and Tokens/Succ. (31,859); Mistral incurs the highest 
latency, consistent with its weaker synthesis in the backbone analysis.

\begin{table*}[t]
\centering
\scriptsize
\setlength{\tabcolsep}{3.5pt}
\renewcommand{\arraystretch}{1.12}
\caption{\textbf{Efficiency analysis on GeoDisaster.} $\uparrow$/$\downarrow$: 
higher/lower is better. Per-trajectory averages: \textit{Agents}, 
\textit{Tools}, \textit{Tokens}. Success-normalized: \textit{Tokens/Succ.}, 
\textit{Latency/Succ.} (s), \textit{Calls/Succ.} Dashes: unavailable or 
undefined (zero successes).}
\label{tab:efficiency_analysis}
\vspace{4pt}
\resizebox{\textwidth}{!}{
\begin{tabular}{l|c|cccccccc}
\toprule
\rowcolor{blue!10}
\textbf{Model} & \textbf{Type}
& \textit{Ans}$\uparrow$
& \textit{Avg. Agents}$\downarrow$
& \textit{Avg. Tools}$\downarrow$
& \textit{Avg. Tokens}$\downarrow$
& \textit{Tokens/Succ.}$\downarrow$
& \textit{Latency (s)}$\downarrow$
& \textit{Latency/Succ.}$\downarrow$
& \textit{Calls/Succ.}$\downarrow$ \\
\midrule
\rowcolor{gray!8}
\multicolumn{10}{l}{\textbf{Single-agent}} \\
Qwen2.5-7B-Inst. & Single 
& 2.53 & 1.00 & 5.33 & 13328 & 526819 & -- & -- & 210.67 \\
\rowcolor{blue!3}
Qwen3-4B-Inst.   & Single 
& 1.16 & 1.00 & 0.23 & 824   & 71072  & -- & -- & 19.83  \\
Llama-3.1-8B-Inst.& Single 
& 0.25 & 1.00 & 0.00 & 379   & 151684 & -- & -- & --     \\
\rowcolor{blue!3}
Mistral-7B-Inst. & Single 
& 0.25 & 1.00 & 0.00 & 379   & 151684 & -- & -- & --     \\
Qwen2.5-VL-7B    & Single 
& 1.50 & 1.00 & 0.00 & 379   & 25280  & -- & -- & --     \\
\midrule
\rowcolor{gray!8}
\multicolumn{10}{l}{\textbf{Multi-agent (no alignment)}} \\
Qwen2.5-7B-Inst. & Multi 
& 22.93 & 2.88 & 6.55 & 28635 & 124880 & 46.22 & 201.57 & 28.57 \\
\rowcolor{blue!3}
Qwen3-4B-Inst.   & Multi 
& 17.91 & 2.44 & 4.70 & 28813 & 160881 & 69.36 & 387.27 & 26.24 \\
Qwen2.5-VL-7B    & Multi 
& 9.21  & 1.00 & 0.00 & 1134  & 12322  & --    & --     & --    \\
\midrule
\rowcolor{gray!8}
\multicolumn{10}{l}{\textbf{Multi-agent (SFT)}} \\
Qwen2.5-7B-Inst. & Multi 
& 82.43 & 3.42 & 4.91 & 28677 & 34790 & 47.51 & 57.64 & 5.96 \\
\rowcolor{blue!3}
Qwen3-4B-Inst.   & Multi 
& 83.29 & 3.21 & 4.58 & 25324 & 30404 & 36.49 & 43.81 & 5.50 \\
Mistral-7B-Inst. & Multi 
& 82.56 & 3.21 & 4.56 & 25241 & 30574 & 35.84 & 43.41 & 5.52 \\
\midrule
\rowcolor{gray!8}
\multicolumn{10}{l}{\textbf{Multi-agent (SFT+GRPO)}} \\
Qwen2.5-7B-Inst. & Multi 
& 90.11 & 3.42 & 4.89 & 28708 & 31859 & 27.64 & 30.67 & 5.43 \\
\rowcolor{blue!3}
Qwen3-4B-Inst.   & Multi 
& 77.28 & 3.40 & 5.06 & 29644 & 38359 & 38.44 & 49.74 & 6.55 \\
Llama-3.1-8B-Inst.& Multi 
& 67.41 & 3.16 & 4.44 & 24127 & 35792 & 38.51 & 57.13 & 6.59 \\
\rowcolor{blue!3}
Mistral-7B-Inst. & Multi 
& 73.97 & 3.21 & 4.49 & 25100 & 33932 & 50.99 & 68.93 & 6.07 \\
\midrule
\rowcolor{gray!8}
\multicolumn{10}{l}{\textbf{Ours}} \\
MAS+SFT          & Multi 
& 82.43 & 3.42 & 4.91 & 28677 & 34790 & 47.51 & 57.64 & 5.96 \\
\rowcolor{teal!14}
MAS+SFT+GRPO     & Multi 
& \textbf{90.11} & \textbf{3.42} & \textbf{4.89} 
& \textbf{28708} & \textbf{31859} & \textbf{27.64} 
& \textbf{30.67} & \textbf{5.43} \\
\bottomrule
\end{tabular}}
\end{table*}

\section{Conclusions}

We presented \textbf{GeoDisaster}, an operational geospatial disaster reasoning benchmark with 2,921 verified instances grounded in executable geospatial workflows, and \textbf{RCEA}, a two-stage alignment protocol for orchestrated multi-agent execution. RCEA combines failure-aware SFT with contract-grounded reinforcement learning to address credit misattribution and reward-scale heterogeneity through dense step-level signals. Experiments show that existing RS-VLMs fall substantially short of operational geo-intelligence, while our aligned framework achieves strong performance across final-answer and trajectory-level metrics on both GeoDisaster and an established remote-sensing agentic benchmark. Failure-wise analysis further shows that SFT improves contract-compliant execution, while GRPO enhances trajectory stability and decision consistency. Overall, the gains across both benchmarks indicate that RCEA supports reliable tool-grounded multi-agent reasoning beyond the proposed dataset. Future work should further improve uncertainty-aware reasoning over multi-source and multi-evidence geospatial settings.

\bibliographystyle{unsrtnat}
\bibliography{ref}

\clearpage \appendix \setcounter{section}{0} \setcounter{figure}{0} \setcounter{table}{0} \setcounter{equation}{0} \renewcommand{\thesection}{S\arabic{section}} \renewcommand{\thefigure}{S\arabic{figure}} \renewcommand{\thetable}{S\arabic{table}} \renewcommand{\theequation}{S\arabic{equation}} \section*{SUPPLEMENTARY MATERIAL} \addcontentsline{toc}{section}{Supplementary Material}

\section{Dataset Details and Statistics}

GeoDisaster contains 2,921 verified instances across five disaster task families and 43 question types. Beyond dataset size, its complexity comes from executable geospatial workflows: instances require an average of 5.00 GT tool calls, 29.10 conversation turns, and 2.52 orchestrator plan steps. The benchmark also includes substantial artifact grounding, with an average of 3.51 initial artifact handles and 16.72 underlying path references per case. Moreover, 42.07\% of the trajectories activate all three specialist agents, indicating that many tasks require coordinated visual, geospatial, and planning reasoning rather than single-step tool use.

\begin{figure}[h]
    \centering
    \includegraphics[width=0.46\linewidth, height= 0.45\linewidth]{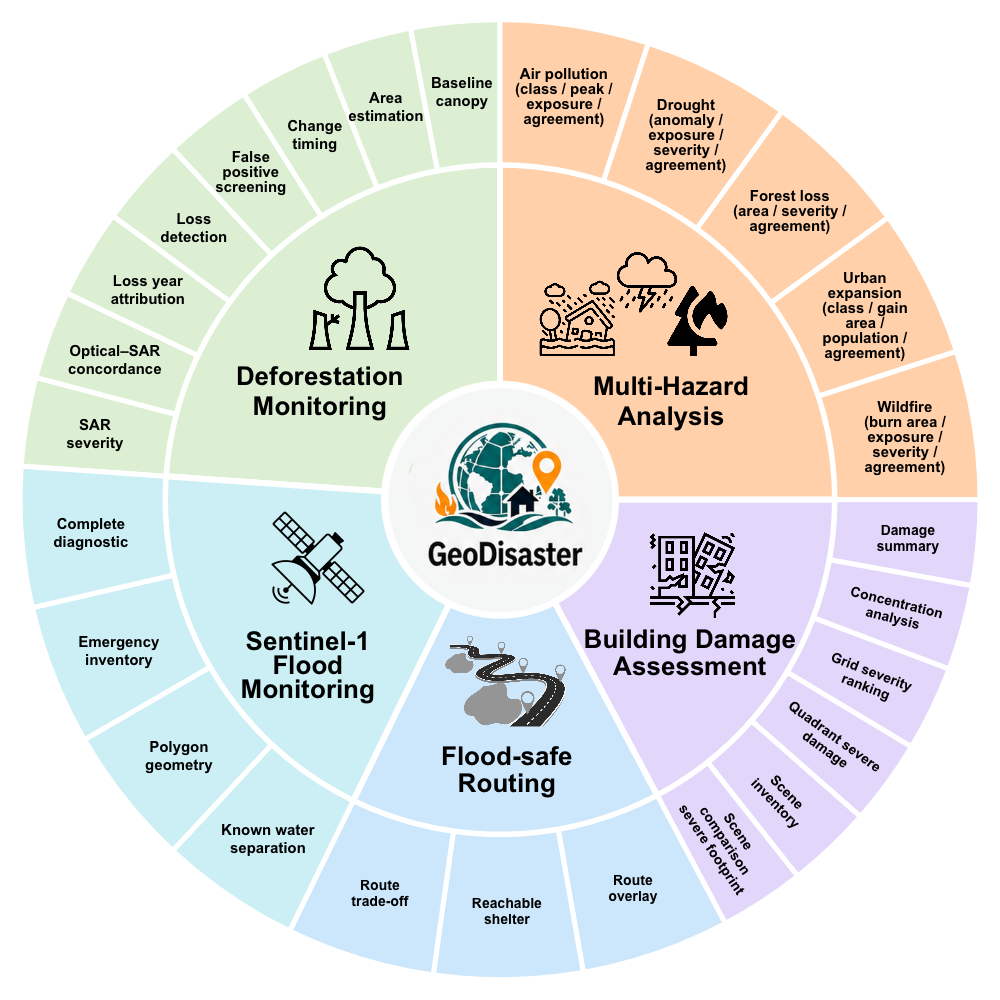}
    \vspace{6pt}
\caption{\footnotesize
GeoDisaster task taxonomy, summarizing the five disaster task families and their associated sub-task categories, highlighting the benchmark's coverage across geo-intelligence tasks.
}
    \label{fig:families}
\end{figure}

\subsection{Task Taxonomy}

GeoDisaster is organized around five disaster-oriented families that capture different forms of operational geo-intelligence: deforestation monitoring, multi-hazard analysis, building-damage assessment, flood-safe routing, and Sentinel-1 flood monitoring. Table~\ref{tab:geodisaster_task_taxonomy} provides the full taxonomy, covering \textit{43} task categories with representative questions that illustrate the expected reasoning behavior in each category. Figure~\ref{fig:families} provides a compact visual summary of these families and their associated sub-task categories, highlighting the benchmark's coverage from monitoring and damage assessment to routing and multi-hazard reasoning.

\begin{table} [!htbp]
\centering
\scriptsize
\setlength{\tabcolsep}{4pt}
\renewcommand{\arraystretch}{1.08}

\resizebox{\textwidth}{!}{
\begin{tabular}{p{0.10\textwidth} p{0.26\textwidth} >{\centering\arraybackslash}p{0.06\textwidth} p{0.56\textwidth}}
\toprule
\rowcolor{gray!12}
\textbf{Family} & \textbf{Task category} & \textbf{N} & \textbf{Representative question} \\
\midrule

\multirow{8}{=}{Deforestation Monitoring}
& Baseline Canopy Density & 48
& What is the baseline tree-cover percentage, and is the canopy dense, open, or sparse? \\
& Change Window Timing & 48
& In which pre/post interval does clearing first become visible? \\
& Deforested Area Estimation & 48
& How many hectares were deforested, and is the loss supported by vegetation-index change? \\
& False Positive Screening & 48
& Is this suspected change true deforestation or a false positive? \\
& Forest Loss Detection & 48
& Did real forest loss occur based on optical, SAR, and Hansen evidence? \\
& Loss Year Attribution & 48
& Which year has the largest forest-loss area, and how many hectares were lost? \\
& Optical--SAR Concordance & 48
& Do optical vegetation change and SAR backscatter change agree? \\
& SAR Severity Proxy & 48
& What is the SAR VH change, and what severity level does it indicate? \\

\midrule

\multirow{20}{=}{Multi-hazard Analysis}
& Air Pollution Class & 20
& What severity class best describes the peak NO$_2$ condition? \\
& Air Pollution Peak & 20
& What is the peak NO$_2$ value and unit? \\
& Drought Anomaly Score & 21
& What is the NDVI anomaly or z-score for this drought location? \\
& Drought Asset Exposure & 21
& How many key assets fall inside the severe-drought footprint? \\
& Drought Severity Class & 21
& What is the dominant drought severity class and fraction? \\
& Drought Source Agreement & 21
& Do precipitation, NDVI anomaly, and soil moisture agree? \\
& Forest Loss Area & 23
& What is the forest-loss area in hectares for the specified year? \\
& Forest Road Proximity & 23
& How far is the clearing centroid from the nearest road? \\
& Forest Severity Class & 23
& What is the dominant forest-loss severity class and fraction? \\
& Forest Source Agreement & 23
& Do Hansen, TMF, and RADD agree on the forest-loss signal? \\
& Pollution Exposure Analysis & 20
& What population or assets are exposed to unhealthy NO$_2$? \\
& Pollution Source Agreement & 20
& Do Sentinel-5P, CAMS, and MODIS AOD agree on the footprint? \\
& Urban Expansion Class & 29
& What is the dominant urban-expansion intensity class? \\
& Urban Gain Area & 29
& How many hectares became newly built-up? \\
& Urban Population Gain & 29
& How much population was added inside the built-up footprint? \\
& Urban Source Agreement & 29
& Do Dynamic World, GHSL, and WorldCover agree on built-up gain? \\
& Wildfire Burned Area & 15
& How many hectares were burned from the burn layers? \\
& Wildfire Exposure Buffer & 15
& How many assets or people fall within the wildfire buffer? \\
& Wildfire Severity Class & 15
& What is the dominant burn-severity class and fraction? \\
& Wildfire Source Agreement & 15
& Do burn products, dNBR, and hotspots agree on the footprint? \\

\midrule

\multirow{7}{=}{Building Damage Assessment}
& Comprehensive Damage Summary & 177
& What is the scene-level damage summary from labels, masks, and severe area? \\
& Damage Concentration Analysis & 121
& Is severe damage concentrated or dispersed across the scene? \\
& Grid Severity Ranking & 177
& Which 4$\times$4 grid cell has the highest damage severity? \\
& Quadrant Severe Damage & 177
& Which quadrant contains the largest severe-damage footprint? \\
& Scene Damage Inventory & 177
& How many buildings fall in each damage class? \\
& Scene Pair Comparison & 55
& Which of two disaster scenes shows greater structural destruction? \\
& Severe Building Footprint & 121
& How many buildings are severely damaged, and what is their area? \\

\midrule

\multirow{3}{=}{Flood-Safe routing}
& Flood Route Overlay & 200
& Can flood extent, blocked roads, shelters, and routes be rendered together? \\
& Nearest Reachable Shelter & 200
& Which shelter is nearest without crossing flooded roads? \\
& Route Tradeoff Selection & 200
& Which routes are fastest, safest, and most balanced? \\

\midrule

\multirow{5}{=}{Sentinel-1 Flood Monitoring}
& Complete Flood Diagnostic & 100
& What is the SAR flood diagnostic, including mask, area, polygons, and grid concentration? \\
& Flood Emergency Inventory & 100
& What are the flood percentage, area, connected regions, and centroid? \\
& Flood Polygon Geometry & 100
& What are the polygon counts, largest polygon, bounding box, and centroid? \\
& Grid Concentration Analysis & 100
& Which 4$\times$4 grid cells are most flooded? \\
& Known Water Separation & 100
& How much predicted flood overlaps known water versus new flood? \\

\bottomrule
\end{tabular}}
\vspace{6pt}
\caption{\textbf{Geo Disaster task taxonomy.} The benchmark contains 43 task categories across five disaster-oriented families. Each category is associated with a representative compact question illustrating the expected reasoning behavior.}
\label{tab:geodisaster_task_taxonomy}
\end{table}

\begin{figure}[h]
    \centering
    \includegraphics[width=0.49\linewidth]{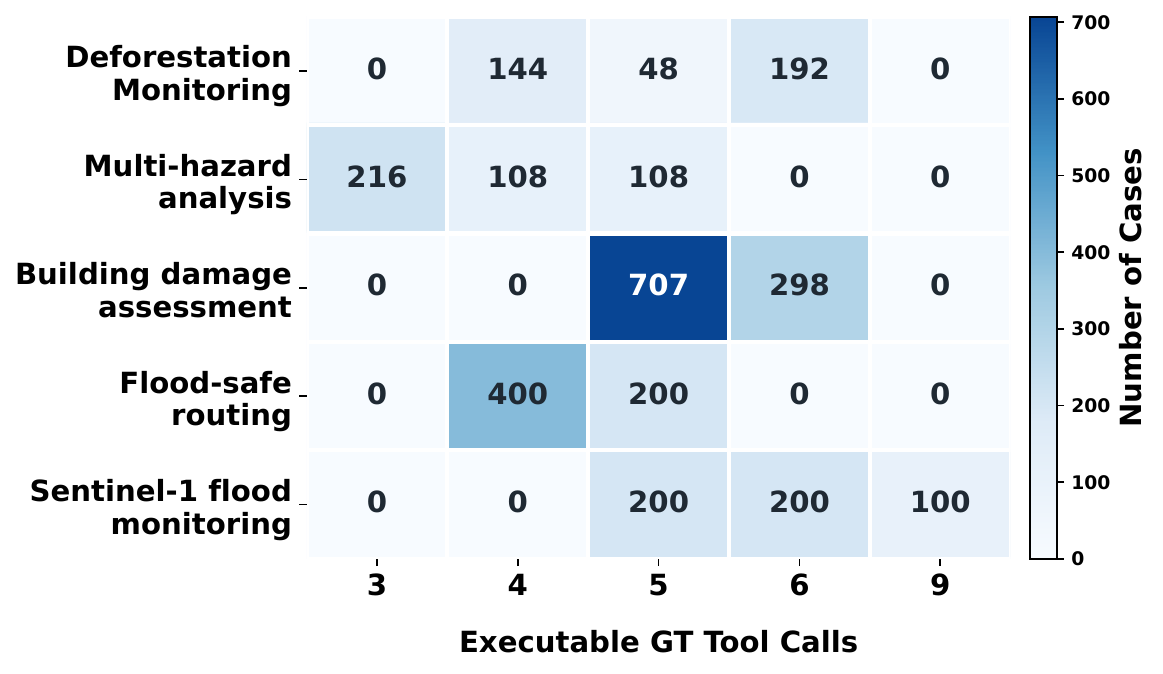}

    \vspace{6pt}
\caption{\footnotesize
Distribution of executable GT tool-chain lengths across GeoDisaster task families. Darker cells indicate more cases, showing how trajectory depth varies across domains.
}
    \label{fig:chains_per_task}
\end{figure}

\noindent \textbf{GT tool-chain lengths.} Beyond category diversity, the taxonomy also reflects heterogeneous execution complexity. As shown in Figure~\ref{fig:chains_per_task}, the distribution of executable GT tool-chain lengths varies substantially across task families. Sentinel-1 flood monitoring and building-damage assessment include longer trajectories due to polygon extraction, grid-level reasoning, spatial concentration analysis, and diagnostic reporting. Multi-hazard analysis contains the largest diversity of question types, covering measurement, severity classification, source agreement, and exposure estimation across several hazard domains. Deforestation monitoring exhibits high artifact density, as tasks require coordinated reasoning over optical, SAR, temporal, and forest-change evidence. Together, this shows that GeoDisaster is diverse not only in task categories, but also in reasoning depth, evidence structure, and tool-chain requirements.

\subsection{Geospatial Tool Suite}
To support tool-grounded disaster reasoning, GeoDisaster defines a compact suite of 18 geospatial tools, summarized in Table~\ref{tab:geodisaster_tools}. These tools abstract dataset-specific operations into reusable reasoning functions covering evidence loading, spatial measurement, temporal and cross-source comparison, impact assessment, route planning, and artifact synthesis. This design serves two purposes. First, it encourages evidence-grounded problem solving, where agents must issue valid tool calls, pass correct handles and arguments, and use intermediate observations rather than relying only on direct answer generation. Second, it enables fine-grained evaluation of where an agent fails: evidence access, spatial computation, tool choice, argument construction, cross-source reasoning, decision planning, or final synthesis. Thus, the tool suite provides both the operational substrate for GeoDisaster tasks and the basis for trajectory-level evaluation beyond final-answer correctness. 

\begin{table}[!htbp]
\centering
\scriptsize
\setlength{\tabcolsep}{3.5pt}
\renewcommand{\arraystretch}{1.12}

\resizebox{\textwidth}{!}{
\begin{tabular}{p{0.20\textwidth} >{\centering\arraybackslash}p{0.13\textwidth} >{\centering\arraybackslash}p{0.06\textwidth} p{0.55\textwidth}}
\toprule
\rowcolor{gray!12}
\textbf{Reasoning role} & \textbf{Tool} & \textbf{Agent} & \textbf{Main function and usage} \\
\midrule

\multirow{2}{=}{Evidence grounding}
& ReadData & GA
& Opens the relevant visible input evidence for a question and prepares a scene-level reference for downstream geospatial analysis. \\
& PrepareScene & GA
& Builds an aligned work context from one or more opened evidence sources, enabling subsequent spatial, visual, or tabular operations. \\

\midrule

\multirow{5}{=}{Spatial measurement and transformation}
& ComputeStats & PA
& Computes numerical summaries such as areas, counts, means, class histograms, footprint statistics, or anomaly scores. \\
& MeasureArea & GA
& Measures the area of selected masks, objects, or extracted regions in the requested units. \\
& ExtractObjects & GA
& Extracts geospatial objects such as forest-loss patches, severe buildings, flood polygons, or connected flood components. \\
& MapFlood & GA
& Produces a flood-mask representation from SAR flood evidence or related flood-monitoring inputs. \\
& DescribeFlood & VRA
& Summarizes flood masks and components using flood percentage, area, polygon geometry, centroid, grid concentration, or water-overlap diagnostics. \\

\midrule

\multirow{4}{=}{Cross-source and temporal reasoning}
& CompareSources & PA
& Compares evidence across modalities or products, such as optical--SAR agreement, multi-source hazard agreement, or known-water overlap. \\
& DetectChange & PA
& Identifies temporal change patterns, dominant disturbance events, or coarse change windows from multi-temporal evidence. \\
& ClassifyChange & PA
& Assigns semantic labels to measured evidence, including canopy state, concordance status, false-positive category, or change severity. \\
& CompareScenes & VRA
& Compares paired scenes or summaries to determine which case has greater damage, risk, or severity. \\

\midrule

\multirow{4}{=}{Impact and risk assessment}
& ScoreHazard & PA
& Converts hazard evidence into severity, concentration, or impact scores used for disaster assessment and prioritization. \\
& EstimateExposure & PA
& Estimates exposed population, roads, facilities, or infrastructure from hazard footprints and contextual GIS layers. \\
& AssessDamage & VRA
& Aggregates building-damage evidence into scene-level, grid-level, quadrant-level, severe-only, or pairwise damage summaries. \\
& RankRegions & PA
& Ranks candidate regions, components, grid cells, quadrants, or routes according to task-specific risk or severity criteria. \\

\midrule

\multirow{2}{=}{Decision and response planning}
& PlanRoutes & PA
& Computes or selects flood-aware shelter routes under avoidance, safety, or reachability objectives. \\
& CompareRoutes & PA
& Compares fastest, safest, and balanced route candidates using travel cost, flood risk, and trade-off criteria. \\

\midrule

\multirow{1}{=}{Artifact synthesis and reporting}
& RenderMap & GA
& Generates map or image artifacts such as flood overlays, route overlays, grid visualizations, or diagnostic figures. \\

\bottomrule
\end{tabular}}
\vspace{6pt}
\caption{\textbf{GeoDisaster geospatial tool suite.} The 18 tools are organized by their role in the operational reasoning workflow, with each tool assigned to the specialist agent responsible for its execution.}
\label{tab:geodisaster_tools}

\end{table}

\subsection{Representative Error Types}

Beyond task-family diversity, GeoDisaster provides trajectory-level supervision for both single-agent and multi-agent execution. Prior agentic systems commonly suffer from structural and semantic failures that are not fully captured by final-answer accuracy, such as invalid planning, wrong tool selection, incorrect argument passing, premature termination, and inconsistent synthesis. By analyzing trajectories across different stages of training, fine-tuning, and alignment, we identify the recurring failure modes summarized in Table~\ref{tab:error_types}.

These failures occur at different levels of the execution process. Planning and delegation errors affect the global workflow before tool execution begins, while tool, argument, and format errors arise during local action generation. Termination, loop, and replanning errors reflect failures in state tracking and control-flow management. This taxonomy motivates our alignment strategy: failure-aware SFT provides corrected recovery actions under the same state and contract, while GRPO encourages contract-consistent execution and penalizes invalid or inefficient trajectories. While our framework significantly reduces these errors quantitatively, several challenges remain, especially for ambiguous visual evidence, invalid geospatial arguments, and premature stopping. These examples suggest that future improvements should focus on uncertainty-aware replanning, stronger argument validation, and better use of multi-temporal evidence.

\begin{table}[!htbp]
\centering
\scriptsize
\setlength{\tabcolsep}{4pt}
\renewcommand{\arraystretch}{1.12}

\resizebox{\textwidth}{!}{
\begin{tabular}{p{0.12\textwidth} p{0.38\textwidth} p{0.23\textwidth} p{0.23\textwidth}}
\toprule
\rowcolor{gray!12}
\textbf{Failure type} & \textbf{Error explanation} & \textbf{Wrong trace} & \textbf{Correct trace} \\
\midrule

\textbf{PlanErr}
& The model fails to decompose the task into required intermediate steps or omits an essential subgoal before execution.
& \texttt{CallAgent(VRA) $\rightarrow$ ClassifyChange}
& \texttt{Plan $\rightarrow$ ReadData $\rightarrow$ DetectChange $\rightarrow$ CompareSources $\rightarrow$ ClassifyChange} \\

\midrule
\textbf{AgentErr}
& The orchestrator assigns a subtask to an unsuitable specialist agent, causing the correct reasoning module to be bypassed.
& \texttt{CallAgent(VRA, MeasureArea)}
& \texttt{CallAgent(GA, MeasureArea)} \\

\midrule
\textbf{ToolErr}
& The selected agent chooses an unsuitable tool for the current subtask, although the required tool exists in the registry.
& \texttt{ToolName = RenderMap}
& \texttt{ToolName = AssessDamage} \\

\midrule
\textbf{ArgErr}
& The agent selects the correct tool but provides invalid, incomplete, or inconsistent arguments.
& \texttt{MeasureArea(mask = null, unit = ha)}
& \texttt{MeasureArea(mask = flood\_mask, unit = ha)} \\

\midrule
\textbf{FormatErr}
& The model violates the required action schema, such as returning free-form text instead of a structured tool call.
& \texttt{``I will compute the flooded area.''}
& \texttt{\{action: MeasureArea, args: \{mask: flood\_mask, unit: ha\}\}} \\

\midrule
\textbf{TermErr}
& The system terminates before all constraints are verified or before sufficient evidence has been collected.
& \texttt{ReadData $\rightarrow$ Terminate}
& \texttt{ReadData $\rightarrow$ MapFlood $\rightarrow$ DescribeFlood $\rightarrow$ Terminate} \\

\midrule
\textbf{LoopErr}
& The system repeats the same action with identical or near-identical arguments without using the returned evidence.
& \texttt{ReadData(x) $\rightarrow$ ReadData(x) $\rightarrow$ ReadData(x)}
& \texttt{ReadData(x) $\rightarrow$ PrepareScene(x) $\rightarrow$ ComputeStats} \\

\midrule
\textbf{ReplanErr}
& The system performs unnecessary replanning after the current evidence is already sufficient to answer the query.
& \texttt{DescribeFlood $\rightarrow$ Plan $\rightarrow$ Plan}
& \texttt{DescribeFlood $\rightarrow$ ComputeStats $\rightarrow$ Terminate} \\

\bottomrule
\end{tabular}}
\vspace{6pt}
\caption{\textbf{Representative error traces.} Common failure modes in GeoDisaster agent execution. Each row gives the failure definition with schematic wrong and correct traces.}
\label{tab:error_types}
\end{table}

\section{Alignment and Implementation Details}

\subsection{Failure-Aware SFT}

We use the error taxonomy in Table~\ref{tab:error_types} to construct failure-aware supervision. Development trajectories are replayed and inspected for recurring structural errors, such as invalid tool calls, stale or missing handles, incorrect arguments, repeated actions, premature termination, and unnecessary replanning. For each failed base-policy trajectory, a teacher model proposes candidate recovery actions conditioned on the task schema, role--tool ownership rules, shared state, and reference artifacts. A candidate is retained only if it passes deterministic validation, including schema correctness, legal role--tool assignment, canonical-handle use, successful tool/state replay, and agreement with the ground-truth answer artifact. Thus, $\mathcal{D}_i$ contains programmatically verifiable recovery actions rather than unchecked teacher outputs, converting observed execution failures into role-specific supervision. As a result, SFT primarily improves handle discipline, valid tool invocation, argument construction, and structured handoff behavior, which explains the large post-SFT gains in Figure ~\ref{fig:task_breakdown} across all task families.

\subsection{Role-Contract GRPO Alignment}
After SFT, GRPO is used to refine trajectory-level decisions that are difficult to capture through imitation alone. The reward combines final task success with role-specific contract satisfaction, robustness, and efficiency terms. Contract checks include schema validity, handle existence, spatial overlap or containment, route--flood intersection, area/count tolerances, and required-field coverage, with borderline spatial cases resolved using dataset-resolution and task-specific tolerance thresholds. Rewards are normalized within each role, so the orchestrator and specialist agents are updated according to their own execution distributions rather than a shared global scale. This is important because ORC, GA, PA, and VRA differ in activation frequency, action type, and failure modes.

\begin{figure}
    \centering
    \includegraphics[width=0.5\linewidth]{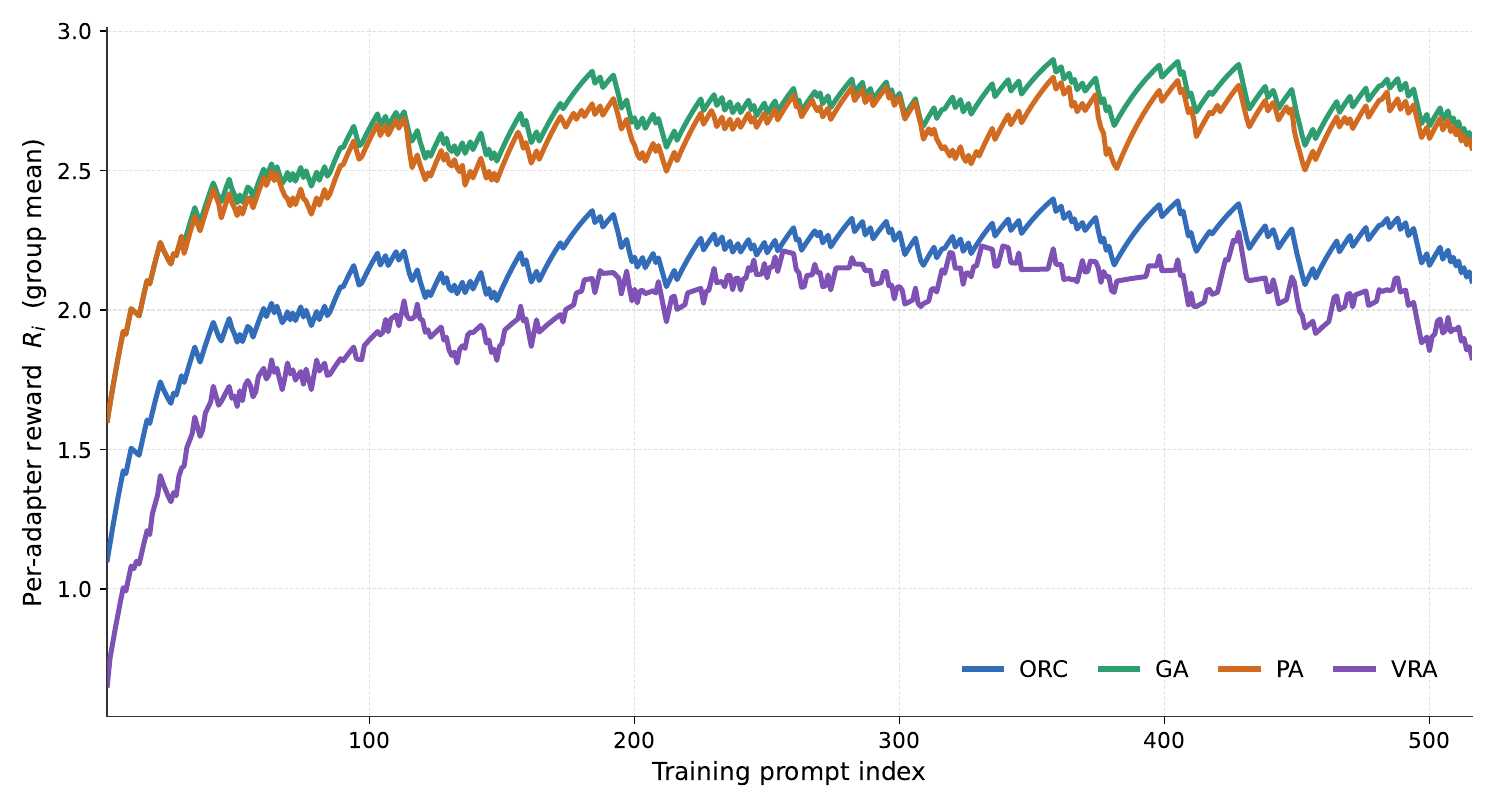}
    \vspace{6pt}
    \caption{\footnotesize
Smoothed role-wise reward curves during GRPO alignment. Rewards increase across all agent roles and stabilize over training, indicating that the alignment objective improves role-specific execution while preserving distinct reward profiles for ORC, GA, PA, and VRA.
}
    \label{fig:grpo1}
\end{figure}

\begin{figure}
    \centering
    \includegraphics[width=0.49\linewidth]{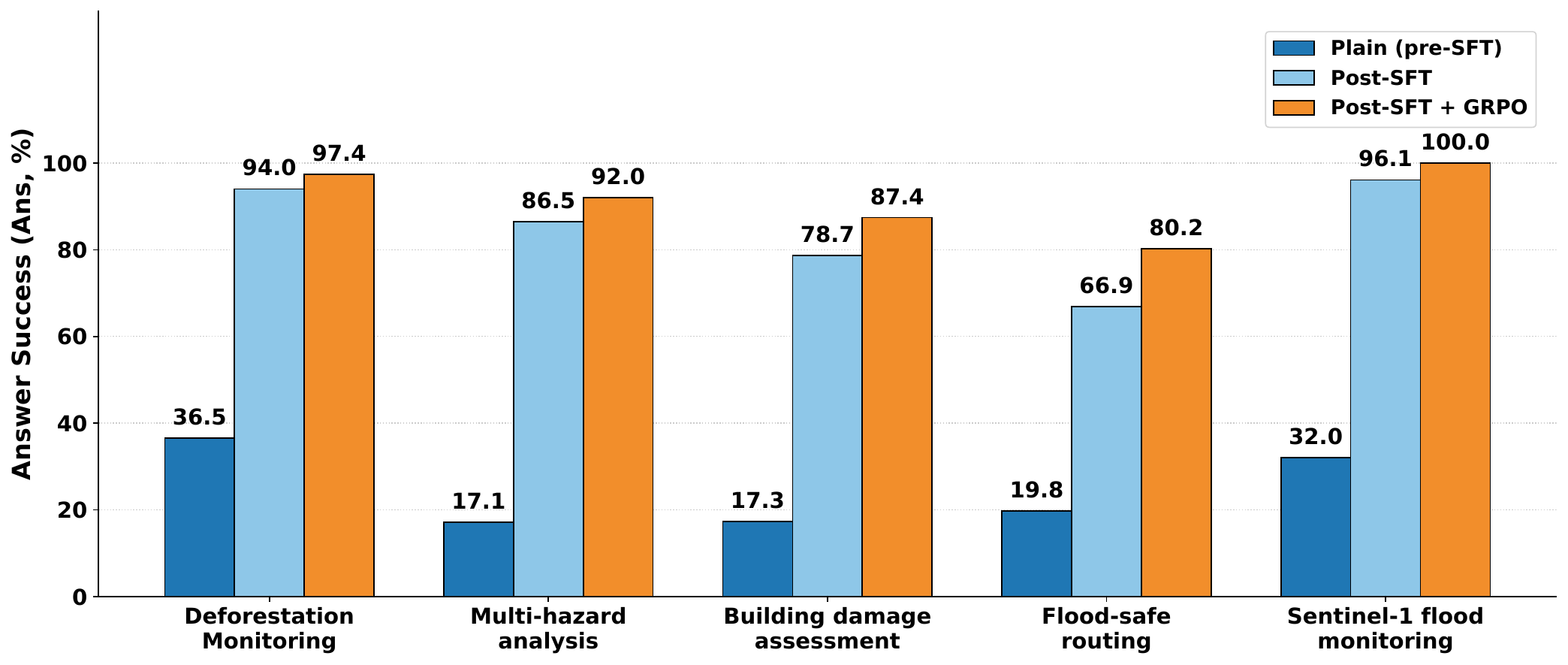}
    \vspace{6pt}
\caption{\footnotesize
Task-wise answer success on GeoDisaster before and after alignment. SFT provides the dominant improvement across all families, while GRPO adds consistent gains, especially on decision-sensitive tasks such as flood-safe routing and building-damage assessment.
}
    \label{fig:task_breakdown}
\end{figure}

Figure~\ref{fig:grpo1} shows that role-wise rewards increase and then stabilize during GRPO alignment, indicating that the policy improves without collapsing the distinct behavior of different agents. GA and PA obtain higher rewards because many tasks involve measurable geospatial operations and analytic decisions, while ORC and VRA show more moderate but stable improvements in coordination and evidence interpretation. The task-wise results in Figure ~\ref{fig:task_breakdown} further show that the vanilla \textit{MAS} remains limited across families, indicating that role decomposition alone does not ensure reliable execution. SFT provides the major improvement by correcting common procedural failures, such as invalid tool use, weak handle discipline, and incomplete response structure. GRPO then adds consistent gains, most clearly in flood-safe routing and building-damage assessment, where tasks involve longer dependent tool chains, ranking decisions, and stricter constraint handling. In contrast, deforestation and Sentinel-1 flood monitoring show smaller post-GRPO gains because many of their errors are already addressed by supervised recovery trajectories. Overall, SFT improves execution validity, while GRPO complements it by refining trajectory-level decisions in tasks where imitation alone is less sufficient.

Overall, while SFT and GRPO substantially improve execution reliability, remaining errors are mainly linked to ambiguous evidence, metric-selection mistakes, and constraint-aware synthesis. This suggests that future work should strengthen uncertainty-aware multi-source geospatial reasoning, especially when modalities provide incomplete, noisy, or partially inconsistent evidence.

\subsection{Training Setup}
Training is implemented in PyTorch/CUDA, while geospatial execution uses \texttt{rasterio}, \texttt{osmnx}, \texttt{earthengine-api}, \texttt{geemap}, and FastAPI-based tool workers. Experiments are run on an AMD EPYC 9654 server with \textit{96} cores / \textit{192} threads, \textit{251} GiB RAM, \textit{119} GiB swap, and \textit{3} NVIDIA RTX PRO 6000 Blackwell Max-Q GPUs, each with approximately \textit{98} GB VRAM. For GRPO alignment, we use \textit{2,064} rollout episodes with group size $K=\textit{4}$, temperature $T=\textit{0.7}$, and top-$p=\textit{0.95}$ sampling. The PPO-style clip is $\epsilon=\textit{0.2}$ and the KL coefficient is $\beta_{\rm KL}=\textit{0.04}$ to keep the policy close to the SFT reference. We train LoRA adapters with rank $r=\textit{32}$, $\alpha=\textit{64}$, learning rate $1\times10^{-6}$, and gradient clipping at \textit{1.0}.

% \begin{figure}
%     \centering
%     \includegraphics[width=0.48\linewidth]{data/domain_answer_success.pdf}
%     \caption{\footnotesize Answer success across GeoDisaster task families before and after alignment.}
%     \label{fig:task_breakdown}
% \end{figure}

\section{Qualitative Examples and Failure Analysis}

\subsection{Alignment Success and Failure Cases}

\begin{figure}
    \centering
    \includegraphics[width=\linewidth]{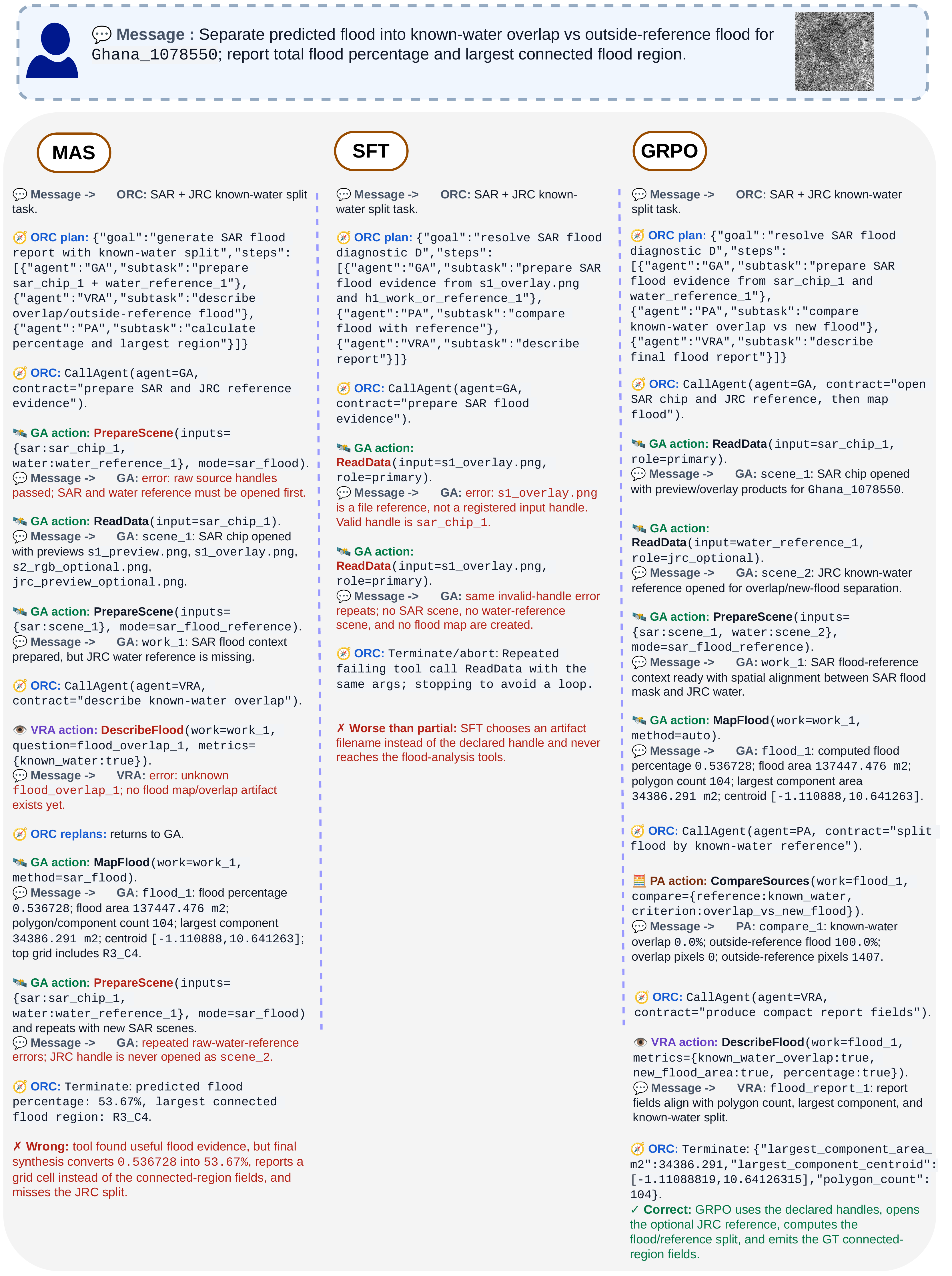}
    \vspace{6pt}
\caption{\scriptsize
Example of successful alignment on a GeoDisaster SAR flood task. The vanilla \textit{MAS} extracts useful intermediate evidence but fails to project it into the final answer. \textit{MAS + SFT} still fails due to an invalid filename/handle, while \textit{MAS + SFT + GRPO} corrects the handle usage and produces a clean completed report.
}
    \label{fig:succeess_e}
\end{figure}

To better understand how the proposed framework behaves on GeoDisaster, we examine selected trajectories that expose both the effect of alignment and the remaining challenges. Figure~\ref{fig:succeess_e} presents a successful alignemt case on SAR flood-monitoring task. The vanilla \textit{MAS} obtains partial flood evidence but fails to convert it into the required structured response, while \textit{MAS + SFT} still suffers from incorrect handle usage. After GRPO, the agent follows the declared evidence handles, opens the SAR and water-reference inputs correctly, performs the required comparison, and produces the requested connected-region report. This example illustrates how the alignment pipeline improves not only final correctness, but also evidence access, tool sequencing, and inter-agent coordination.

\begin{figure}
    \centering
    \includegraphics[width=\linewidth]{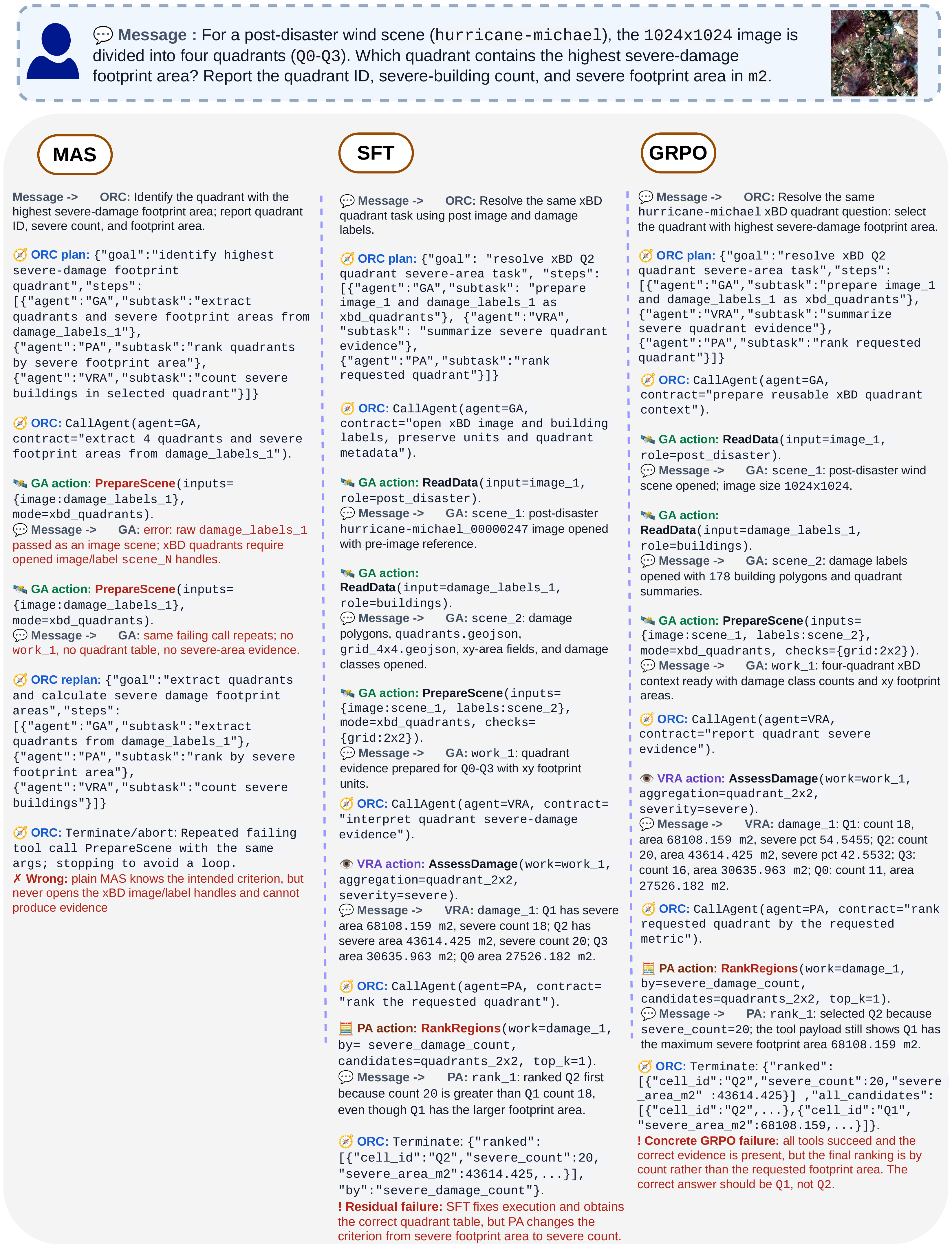}
    \vspace{6pt}
\caption{\scriptsize
Failure-case comparison on a GeoDisaster damage-assessment task. The plain MAS fails to access valid evidence handles. SFT fixes evidence extraction but still ranks by building count instead of footprint area. GRPO preserves the improved execution, but the final answer retains the same metric-selection error.
}
    \label{fig:fail_e}
\end{figure}

Figure~\ref{fig:fail_e} shows a contrasting damage-assessment case where the framework partially fails after alignment. Here, SFT and GRPO both improve the execution process by recovering valid image and label evidence, preparing the quadrant context, and extracting the relevant damage statistics. However, the final decision uses the wrong ranking criterion, selecting the quadrant with the highest severe-building count rather than the largest severe-damage footprint area requested in the query. This case shows that alignment substantially improves procedural reliability, but criterion-sensitive synthesis remains difficult when multiple valid metrics are available.

\subsection{Qualitative Results}

\begin{figure}
    \centering
    \includegraphics[width=0.95\linewidth]{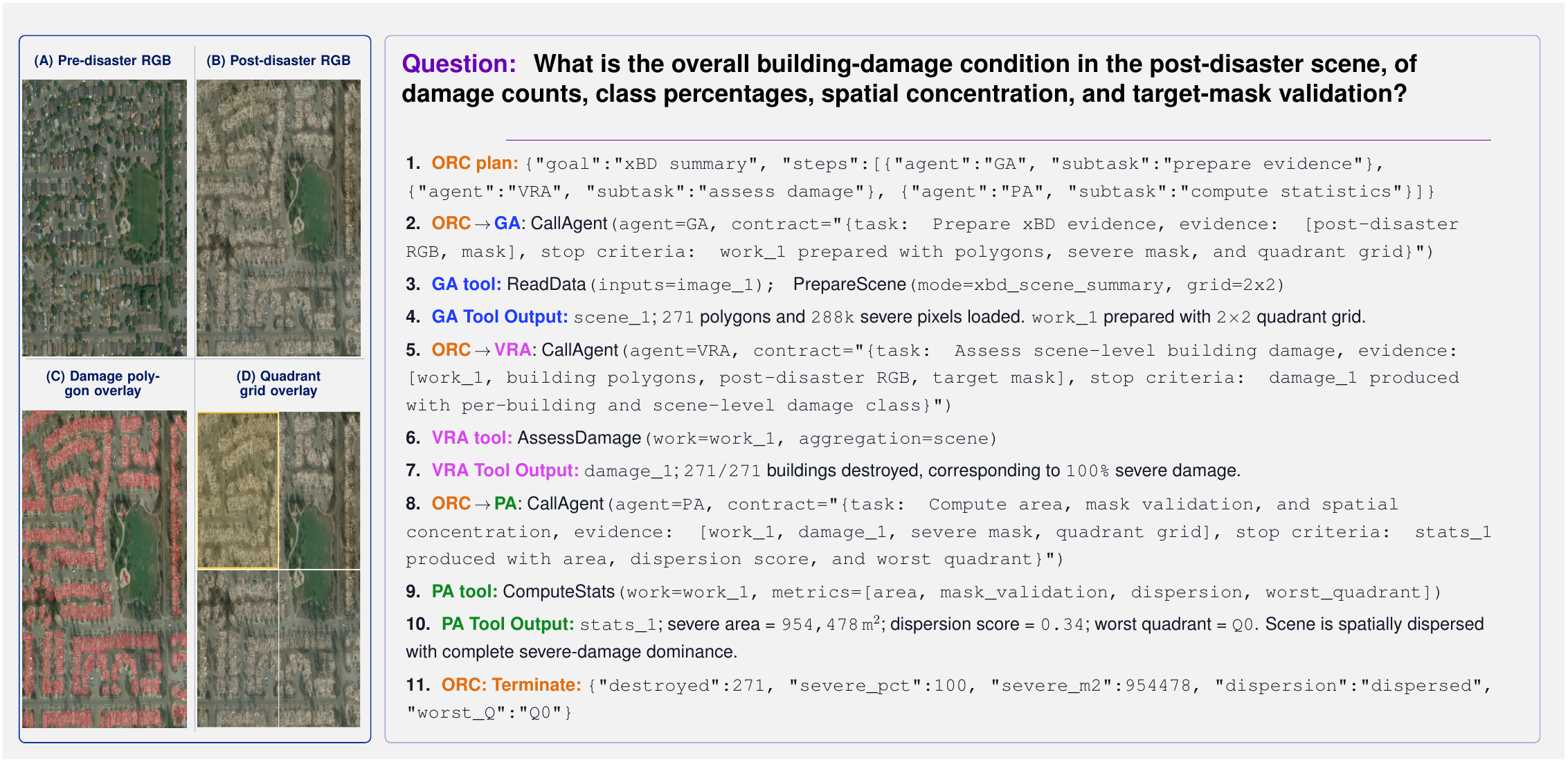}
    \vspace{6pt}
    \caption{\footnotesize
Example GeoDisaster for scene-level building-damage assessment. The task combines pre/post imagery, damage polygons, quadrant overlays, and role-specialized tool calls to produce a structured damage summary with counts, severe-area estimates, spatial dispersion, and mask validation.
}
    \label{fig:damage}
\end{figure}

\begin{figure}
    \centering
    \includegraphics[width=0.95\linewidth]{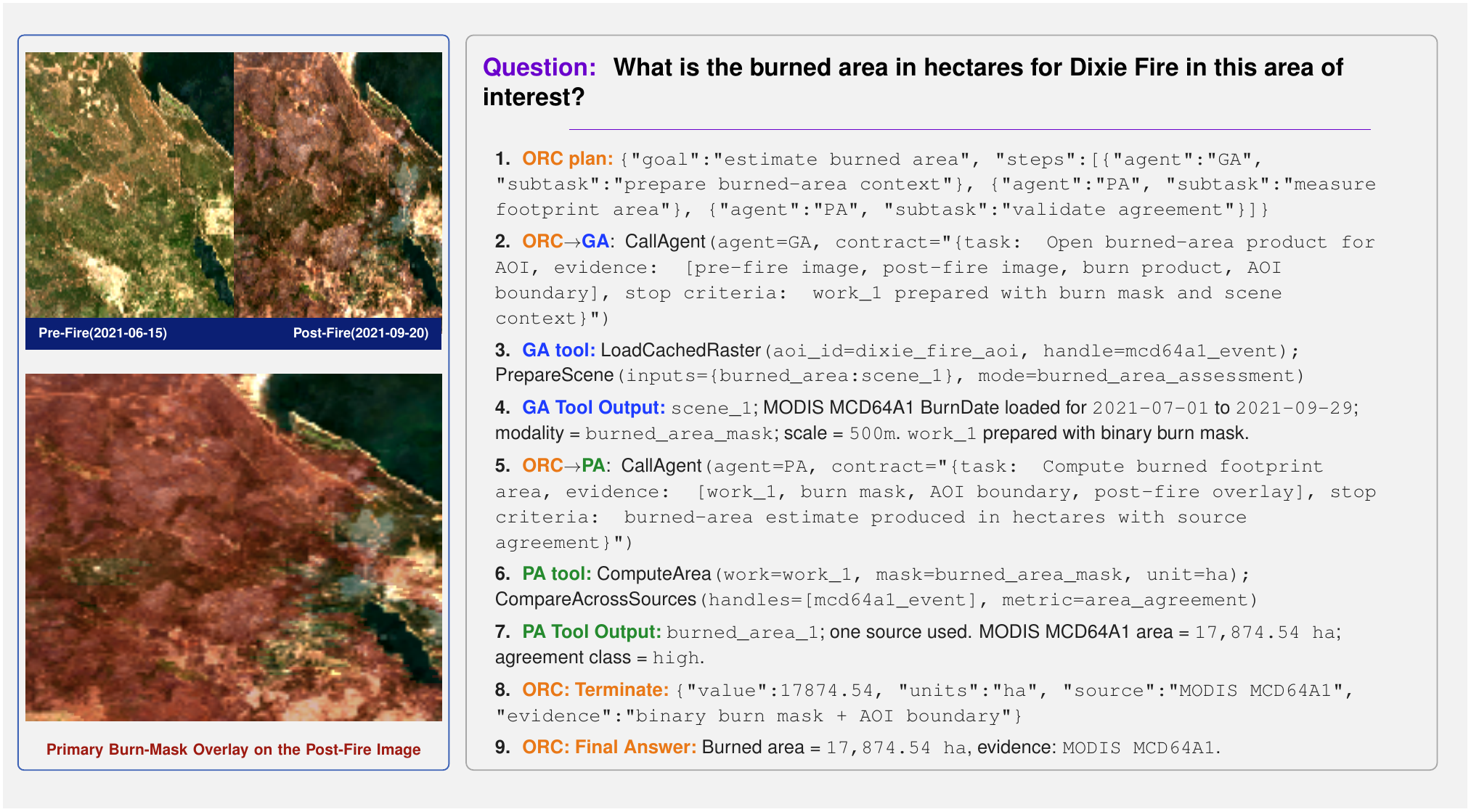}
    \vspace{6pt}
    \caption{\footnotesize
Example GeoDisaster for wildfire burned-area estimation. The task uses pre/post fire imagery and a burn-mask product to compute the burned footprint area, with role-specialized tool calls for evidence loading, area measurement, and final grounded reporting.
}
    \label{fig:wildfire}
\end{figure}

We provide representative GeoDisaster task examples from two task families to illustrate how the benchmark captures full conversational trajectories among agents rather than only final answers.  Figure~\ref{fig:damage}, shows a building-damage assessment task, where all agents contribute to the successful completion. The orchestrator decomposes the query into subtasks of evidence preparation, damage interpretation, and spatial summary steps. The trajectory shows how agents use pre/post imagery, damage labels, and spatial overlays to produce a grounded scene-level report with damage counts, severe-area estimates, spatial dispersion, and mask-based validation. Figure~\ref{fig:wildfire} presents an example of  wildfire analysis, where the agents estimate burned area from pre/post fire evidence and a burn-mask product. The conversation illustrates the flow from evidence loading to area measurement and final reporting, showing how specialist agents contribute intermediate observations that are consolidated into a grounded answer. Together, these examples show that GeoDisaster evaluates the complete reasoning process, including orchestration, tool execution, evidence grounding, and final synthesis.

\subsection{LLM-as-Judge Evaluation Protocol}

\begin{figure}
    \centering
    \includegraphics[width=0.99\linewidth]{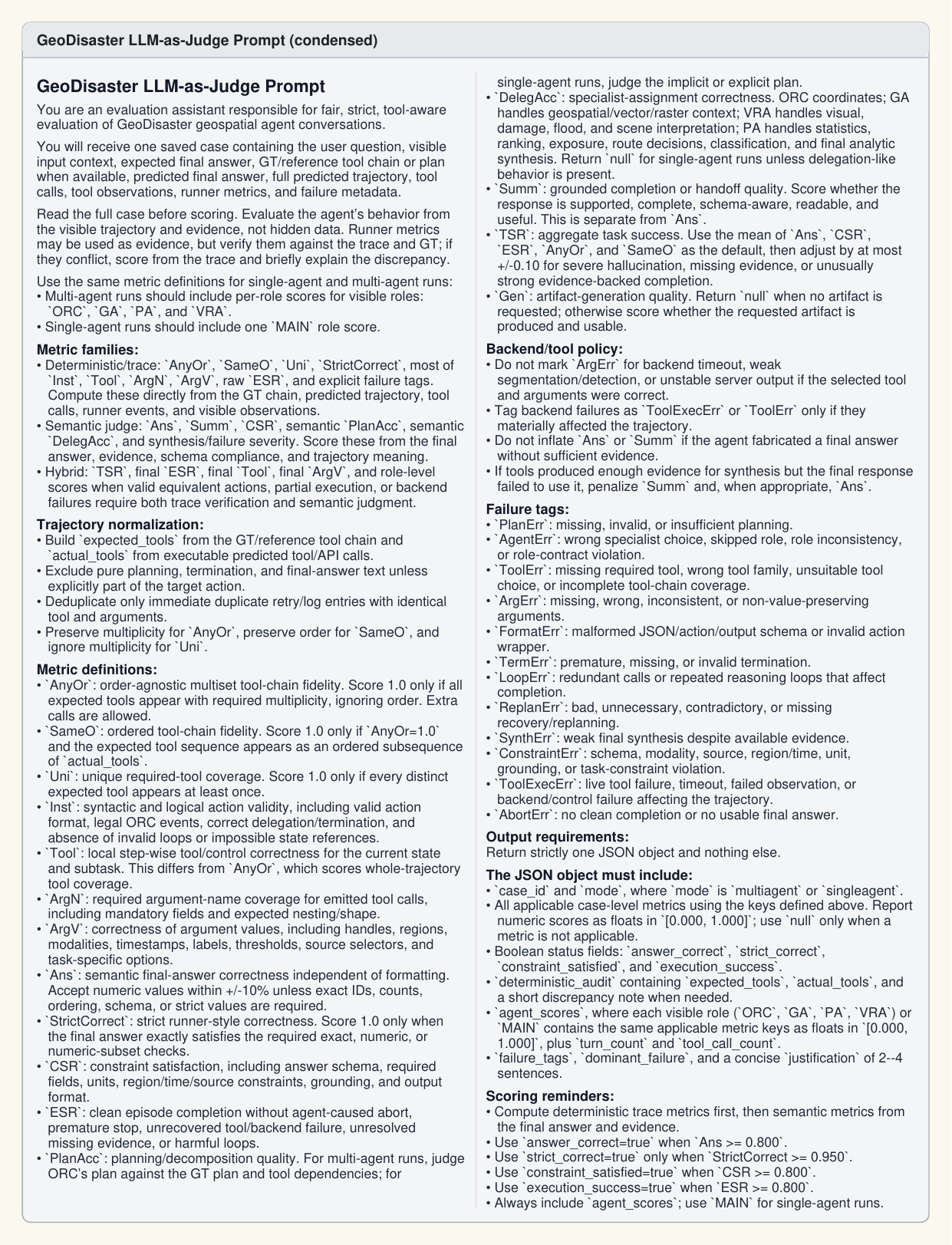}
    \vspace{6pt}
\caption{\scriptsize
Condensed LLM-as-judge prompt used for GeoDisaster evaluation. The prompt defines deterministic, semantic, and hybrid metric families; specifies scoring rules for tool-chain fidelity, final-answer correctness, constraint satisfaction, execution success, role-level agent scores, and failure tags; and enforces structured JSON output for reproducible aggregation.
}
    \label{fig:llm_judge}
\end{figure}

We use GPT-5.5 as an LLM-assisted evaluator for GeoDisaster trajectories. The evaluator is given the complete saved case, including the user question, visible input context, expected answer, reference tool chain or plan, predicted final answer, full predicted trajectory, tool calls, tool observations, runner metrics, and failure metadata. Importantly, the judge is instructed to inspect the \textit{full execution trace} and separate \textbf{deterministic} trace-based metrics from \textbf{semantic} and \textbf{hybrid} judgments. To check reliability, we manually reviewed a subset of \textit{50} evaluated cases and revised the prompt until the produced metric judgments were consistent with human inspection. Figure~\ref{fig:llm_judge} shows condensed LLM-as-judge prompt used for our evaluations.

The evaluation uses two main metric groups. \textbf{End-to-end metrics} measure whether the full trajectory reaches the intended task outcome. \textit{ToolAnyOr}, \textit{ToolSameO}, and \textit{ToolUni} evaluate tool-chain fidelity under order-agnostic multiset matching, exact ordered matching, and unique-tool coverage, respectively. \textit{TSR} summarizes overall task success by combining final-answer quality, episode completion, tool coverage, and tool-order fidelity. \textit{Ans} measures task-level correctness for non-generative outputs, while \textit{Gen} measures artifact-generation quality when the task requires a map, image, or visual output. \textit{CSR} measures whether the final answer satisfies schema, grounding, units, source, temporal, and task-specific constraints. \textit{ESR} captures clean episode completion, penalizing premature stopping, unresolved failures, harmful loops, weak synthesis, and invalid plans. For multi-agent settings, \textit{PlanAcc} measures the quality of the orchestrator's decomposition, and \textit{DelegAcc} evaluates whether subtasks are assigned to the appropriate specialist agents.

\textbf{Step-wise execution metrics} evaluate the internal quality of the trajectory. \textit{Inst} measures whether actions are syntactically valid, logically executable, and consistent with the required control format. \textit{Tool} measures local tool or control-action correctness at each step, which differs from whole-trajectory tool-chain fidelity. \textit{ArgN} measures the presence of required tool parameters, while \textit{ArgV} measures the correctness of parameter values, including handles, regions, modalities, timestamps, source selectors, thresholds, and task-specific options. \textit{Summ} measures the quality of grounded intermediate completion or handoff, including whether the produced summary is concrete, evidence-supported, schema-aware, and useful for downstream reasoning. It is therefore distinct from \textit{Ans}, which measures final-answer correctness against the expected answer.

The prompt further distinguishes three scoring modes. \textit{Deterministic/trace metrics} are computed directly from the reference chain, predicted tool calls, runner events, and visible observations; these include tool-chain fidelity, strict correctness, most instruction/tool/argument checks, raw episode resolution, and explicit failure tags. \textit{Semantic metrics}, such as \textit{Ans}, \textit{Summ}, \textit{CSR}, \textit{PlanAcc}, and \textit{DelegAcc}, require interpretation of the final answer, evidence use, schema compliance, and trajectory meaning. \textit{Hybrid metrics}, such as \textit{TSR}, final \textit{ESR}, final \textit{Tool}, final \textit{ArgV}, and role-level scores, combine trace verification with semantic judgment when valid equivalent actions, partial execution, or backend/tool failures require contextual interpretation. The judge returns a structured JSON object containing case-level metrics, role-level scores for ORC, GA, PA, and VRA or MAIN for single-agent runs, deterministic audit fields, failure tags, a dominant failure type, and a concise justification.

\vspace{-10pt}

\end{document}